\documentclass[twocolumn,10pt]{IEEEtran}
\usepackage[mathscr]{eucal}
\usepackage{ifpdf}
\usepackage{cite}
\usepackage{graphicx}
\usepackage[cmex10]{amsmath}
\usepackage{amssymb}
\usepackage{amsmath}
\usepackage{placeins}
\DeclareMathOperator*{\argmax}{arg\,max}
\DeclareMathOperator*{\argmin}{arg\,min}
\usepackage{algorithmic}
\usepackage{units}
\usepackage{setspace}
\usepackage{algorithm}
\usepackage{array}
\usepackage{booktabs}
\usepackage{amsthm}
\usepackage{multirow}
\usepackage{enumerate}
\usepackage{color}
\usepackage{mathtools}
\usepackage{caption}
\usepackage{subcaption}
\usepackage{tabularx}
\usepackage{tcolorbox}
\usepackage{supertabular}
\usepackage{tikz}
\usepackage{bm}
\usepackage{xcolor,colortbl}
\usepackage{titlesec}
\usepackage{enumitem}

\usetikzlibrary{arrows.meta, positioning}

\usepackage{pifont}
\newcommand{\cmark}{\ding{51}}%
\newcommand{\xmark}{\ding{55}}%

\newcounter{assume}

\newcommand{\resl}[1]{}
\newcommand{\tosl}[1]{}
\newcommand{\cosl}[1]{}

\definecolor{mygreen}{RGB}{143, 188, 187}
\definecolor{green1}{RGB}{147, 196, 125}
\definecolor{red1}{RGB}{213,88,88}
\begin{document}
\title{Single and Multi-Agent Deep Reinforcement Learning for AI-Enabled Wireless Networks: A Tutorial}
\author{
\IEEEauthorblockN{Amal Feriani and Ekram Hossain, \textit{Fellow, IEEE}}
}
\maketitle
%
\begin{abstract}
Deep Reinforcement Learning (DRL) has recently witnessed significant advances that have led to multiple successes in solving sequential decision-making problems in various domains, particularly in wireless communications. The future sixth-generation (6G) networks are expected to provide scalable, low-latency, ultra-reliable services empowered by the application of data-driven Artificial Intelligence (AI). The key enabling technologies of future 6G networks, such as intelligent meta-surfaces, aerial networks, and AI at the edge, involve more than one agent which motivates the importance of multi-agent learning techniques. Furthermore, cooperation is central to establishing self-organizing, self-sustaining, and decentralized networks.  In this context, this tutorial focuses on the role of DRL with an emphasis on deep Multi-Agent Reinforcement Learning (MARL) for AI-enabled 6G networks. The first part of this paper will present a clear overview of the mathematical frameworks for single-agent RL and MARL. The main idea of this work is to motivate the application of RL beyond the model-free perspective which was extensively adopted in recent years. Thus, we provide a selective description of RL algorithms such as Model-Based RL (MBRL) and cooperative MARL and we highlight their potential applications in 6G wireless networks. Finally, we overview the state-of-the-art of MARL in fields such as Mobile Edge Computing (MEC), Unmanned Aerial Vehicles (UAV) networks, and cell-free massive MIMO, and identify promising future research directions. We expect this tutorial to stimulate more research endeavors to build scalable and decentralized systems based on MARL.

\let\thefootnote\relax\footnotetext{The authors are with the Department of Electrical 
and Computer Engineering, University of Manitoba, Winnipeg, MB, Canada (e-mails: 
feriania@myumanitoba.ca, ekram.hossain@umanitoba.ca). The work was supported by a Discovery Grant from the Natural Sciences and Engineering Research Council of Canada (NSERC).}
\end{abstract}

\section{Introduction}
\label{Sec:Intro}

The evolution of wireless networks such as the fifth-generation (5G) and beyond 5G (also referred to as 6G) networks is driven by a huge increase of connected devices, the growing demand for high data rate applications, and the convergence between communications and computing. Intelligent signal processing along with AI-driven super-intelligent radio access and network control will be among the key technologies in future wireless networks to achieve scalability, context-awareness, and energy efficiency along with massive capacity and connectivity. 5G wireless networks are expected to provide Ultra-Reliable, Low Latency Communications (URLLC),  Enhanced Mobile Broadband (eMBB), and massive Machine-Type Communications (mMTC) to enable the deployment of Internet-of-Things (IoT) services such as e-Health, smart cities, extended reality, etc. Meanwhile, several research initiatives \cite{saad20196G}, \cite{akyildiz20206g}, \cite{bariah20206G} proposed road-maps or visions for 6G networks to overcome the limitations of  5G technologies. All these works envision that 6G networks will rely on autonomous systems providing scalable, reliable, and secure services. For instance, the transition from 5G to 6G will introduce new key technologies such as TeraHertz (THz) and optical wireless communications (e.g. visible light communications), intelligent metasurface-aided wireless communications, aerial networks, and Multi-access Mobile Edge Computing (MEC).

The recent success of AI techniques, namely Machine Learning (ML) and Deep Learning (DL), has spurred the adoption of a \emph{learning} perspective to solve wireless control and management problems. For instance, Deep Neural Networks (DNN) are universal approximators able to estimate any function thus they can approximate optimal solutions for complex tasks. DNNs can be used in three different ML settings: supervised ML, unsupervised ML, and Reinforcement Learning (RL). Although supervised ML methods showed promising results in several wireless problems such as channel estimation and IRS joint beamforming and shift optimization, they require the availability of a large amount of {\em a priori labeled} training and testing data which is difficult to obtain for real-life scenarios. This motivates RL techniques that rely on ``trial and error" to solve sequential decision-making problems. In RL, a learning agent interacts with an environment by choosing an action and observing the system's next state and an immediate reward. Naturally, the agent seeks to find optimal actions to maximize its rewards. To do so, two approaches can be adopted: model-free or model-based. Model-based RL (MBRL) assumes that the agent knows the system dynamics that is how the system transits from one state to another one and how rewards are generated. This approach is not always feasible, especially in complex systems where the agent has restricted to no knowledge about its environment. This motivates model-free techniques where an agent learns optimal strategies without any knowledge about the system. As a generalization of the single-agent RL setting, multi-agent RL seeks to solve decision-making problems involving more than one agent. 

\begin{figure}[ht]
     \centering
         \centering
         \includegraphics[scale=0.5]{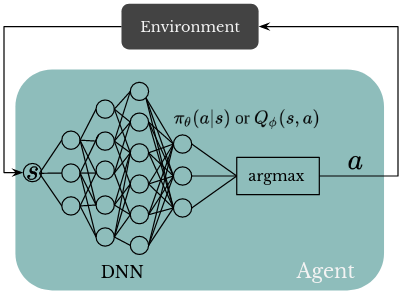}
         \caption{Representation of a DRL framework.}
         \label{fig:DRL}
\end{figure}

In RL, we differentiate between two key steps: \emph{training} and \emph{inference}. During the training phase, the agent interacts with the environment to collect experience. The environment is often a simulator mimicking the real-world system since it is expensive to directly interact with the real system. These collected experiences constitute the training dataset of the RL agent and will be used to learn the optimal decision-making rule. In Deep RL (DRL), DNNs are used to approximate the agent's optimal strategy or policy and/or its optimal utility function (see Figure \ref{fig:DRL}). In this case, given the system's current state, the DNN learns to predict either a distribution over actions or the action expected reward $Q(s, a)$. Therefore, the agent chooses its next action as the one that has either the highest probability or the highest expected reward. After receiving the reward from the environment, the DNN parameters are updated accordingly. The generalization power of DNNs enables solving high-dimensional problems with continuous or combinatorial state spaces. In the context of wireless communications, DRL is advantageous compared to the traditional optimization methods thanks to their \emph{real-time inference}. However, the training phase of DNNs requires a considerable amount of computation power which necessitates the use of GPUs and high-performance CPU clusters. The most popular DL frameworks are \emph{Tensorflow} \cite{abadi2016tensorflow} and \emph{pytorch} \cite{paszke2019pytorch}. Once the training is complete, the agent can make decisions in real-time which is a considerable advantage compared to traditional optimization problems. To accelerate the inference of DNNs further, different libraries implement sophisticated compression techniques such as quantization to fasten the execution of DNNs on mobile or edge devices.

\subsection{Scope of this Tutorial}

In this work, we emphasize the role of DRL in future 6G networks. In particular, our objective is to discuss several DRL learning frameworks to advance the current state-of-the-art and accommodate the requirements of 6G networks. First, we overview single-agent RL methods and shed light on MBRL techniques. Although MBRL has received less interest, it can show considerable advantage compared to model-free algorithms. MBRL consists in learning a model representing the environment dynamics and utilize the learned model to compute an optimal policy. The main advantage of having a model of the environment is the ability to plan ahead which makes these methods more sample-efficient. In addition, MBRL is more robust to changes in the environment dynamics or rewards and has better exploratory behaviors. Recent progress in MBRL, especially for robotics, has shown that MBRL can be more efficient than MFRL. However, for most applications, it is challenging to learn an accurate model of the world. For this reason, model-free algorithms are preferred. However, in the second part of the tutorial, we argue that single-agent RL is not sufficient to model scalable and self-organizing systems often containing a considerable number of interconnected agents. This claim is justified since single-agent RL algorithms learn a decision-making rule for one entity without considering the existence of other entities that can impact its behavior. Thus, we will study extensions of both model-free and model-based single-agent approaches to multi-agent decision-making.

MARL is the generalization of single-agent RL that enables a set of agents to learn optimal policies using interactions with the environment and each other. Thus, MARL does not ignore the presence of the other agents during the learning process which makes it a harder problem. Essentially, several challenges arise in the multi-agent case such as i) {\em non-stationarity} of the system due to the agents simultaneously changing their behaviors; ii) {\em scalability} issue since the joint action space grows exponentially with the number of agents; iii) {\em partial observability} problem arising often in real-world applications where agents have access only to partial information of system; iv) privacy and security are also core challenges of the deployment of MARL systems in real-world scenarios. More details on these issues will be presented in Section \ref{sec:MARLchallenges}. 

MARL is generally formulated as a Markov Game (MG) or also called a Stochastic Game (SG). MG generalizes Markov Decision Processes (MDPs) used to model single-agent RL problems and repeated games in game theory literature. In repeated games, the same players repeatedly play a given game called stage game. Thus, repeated games consider a stateless static environment, and the agents' utilities are only impacted by the interactions between agents. This is a crucial limitation of normal-form game theory frameworks to model multi-agent problems. MG remedies this shortcoming by considering a dynamic environment impacting the agents' rewards. MGs can be classified into three families \emph{fully cooperative}, \emph{fully competitive} or \emph{mixed}. Fully cooperative scenarios assume that the agents have the same utility or reward function whereas fully competitive settings involve agents with opposite goals often known as \emph{zero-sum} games. The mixed setting covers the general case where no restriction on the rewards is considered. This is also referred to as \emph{general-sum} games. In this paper, we focus on \emph{fully cooperative MARL} and consider MGs as the mathematical formalism to model such problems. However, MGs handle only problems with \emph{full observability} or \emph{perfect information}. Other extensions to model partial observability will be discussed as well.

Because fully cooperative agents share the same reward function, they are obliged to choose optimal \emph{joint} actions. In this context, \emph{coordination} between agents is crucial in selecting optimal joint strategies. To illustrate the importance of coordination, we consider the example from \cite{boutilier1996planning}. Let us examine a scenario with two agents at a given state of the environment where they can choose between two actions $a_1$ and $a_2$. We assume that the joint actions $(a_1, a_1)$ and $(a_2, a_2)$ are both optimal joint actions (i.e. yield maximum reward). Without coordination, the first agent can choose $a_1$ and the second agent can pick $a_2$ which result in non-optimal joint action $(a_1, a_2)$. Thus, although both agents individually choose decisions induced by one of the optimal joint actions, the resultant joint action is not optimal. This is why coordination is important to make sure agents choose the \emph{same} joint action profile. The most straightforward way to establish collaboration is to assume the existence of a central unit to collect information from the agents, find the optimal joint action, and communicate it back to the agents. In practice, however, this assumption is either costly or infeasible. Ideally, the agents learn to adapt their own policies with limited to no communication while ensuring coordination. This motivates the \emph{decentralized} approaches in solving MARL problems.

Different approaches can be adopted to solve fully cooperative tasks depending on the information used by the agents for learning. \emph{Independent Learners} (IL) is a fully decentralized scheme where agents have access to their local information only and independently optimize their policies to maximize their returns. However, this approach rules out the cooperation between agents and ignores the non-stationarity issue. Classic MARL approaches adopted \emph{Nash Equilibrium} (NE) as a solution concept. Nevertheless, as outlined by \cite{shoham2003MAL}, solving multi-agent scenarios using NE is problematic for several reasons. First, finding Nash equilibria is challenging and proved to be PPAD-complete for two-player general-sum games \cite{chen2009NEComplexity}. Furthermore, in the case of multiple equilibria, it is questionable how the agents will converge toward the same equilibrium. This encourages other solution concepts to solve cooperative multi-agent problems. In this tutorial, we will examine DRL techniques for learning optimal joint policies. We distinguish between three sub-fields of deep MARL: \emph{learned cooperation}, \emph{emergent communication}, and \emph{networked agents}. Learned cooperation seeks to learn cooperative behaviors without any communication means. If communication is permitted, the agents can either learn an efficient communication protocol as in the emergent communication field or assume the existence of a communication structure like in the networked agents framework. The latter is advantageous since it considers heterogeneous agents with different goals cooperating to maximize the team average return. On the contrary, the other learning methods assume homogeneous agents with the same reward function.

\begin{figure}[ht]
     \centering
         \centering
         \includegraphics[scale=0.35]{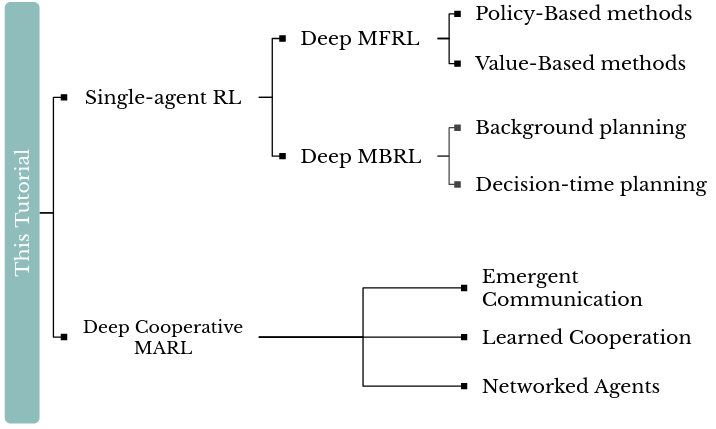}
         \caption{Scope of the tutorial.}
         \label{fig:tutoTaxonomy}
\end{figure}

The scope of this tutorial is summarized in Figure \ref{fig:tutoTaxonomy}. The core motivation of this work is to provide a tutorial on the tools enabling the design of multi-agent algorithms in a decentralized fashion. Decentralized MARL is an active area of research. The expertise of the wireless communication community can boost the progress of this field via designing efficient means of communication between agents. {\em Other multi-agent learning strategies such as \emph{federated learning} and \emph{distributed RL} are not covered in this tutorial}. Federated Learning aims to learn a common global model in a distributed fashion while maintaining the privacy of the learning agents. This is different from cooperative MARL where the agents seek to learn their own policies without a central coordinator. In addition, distributed RL aims to accelerate the training of DRL algorithms using sophisticated distributed computing techniques. From now on, we will use ``distributed” and ``decentralized” interchangeably; unless otherwise stated, they both refer to decentralized agents without a central controller.


\subsection{Existing Surveys and Tutorials}
DRL has become the cornerstone of a plethora of algorithms for intelligently solving complex tasks in multiple areas such as MEC, aerial networks, etc. Several efforts have focused on summarizing these contributions. For example, \cite{luong2019Survey} provides a comprehensive overview of single-agent DRL algorithms for communications and networking problems such as dynamic network access, data rate control, wireless caching, data offloading, network security, and connectivity preservation. \cite{qian2019surveyDRL} addresses the recent DRL contributions in MEC, software-defined networking, and network virtualization in 5G.
Other surveys focus on specific applications of DRL such as autonomous Internet-of-Things \cite{lei2020AIOTDRL}, resource management for 5G heterogeneous networks \cite{lee2019survey5GHetNets}, and mobile edge caching \cite{zhu2018DRLcaching}. \textbf{Most of these works are restricted to model-free single-agent DRL methods. In addition to model-free DRL, our work focuses on other less explored RL techniques such as MBRL for single-agent settings and cooperative MARL for multiple agent scenarios.}

More recently,  the role of DRL in 6G networks has been a topic of increasing interest. A white paper \cite{ali20206g} on ML for 6G networks overviews the key ML enabling techniques such as RL and federated learning and discusses the potential applications in different network layers in addition to highlighting several open research directions. In \cite{tang20196GUAV}, the authors reviewed the applications of ML, particularly DRL, in vehicular networks with a discussion on the role of AI in the future 6G vehicular networks.
Furthermore, \cite{she2020tutorial6GURLLC} considers the role of DL and DRL for URLLC communications in future 6G networks. Similarly, these contributions mostly focus on single-agent DRL for future 6G networks with limited exposure to decentralized learning. 

In this paper, we argue that single-agent RL methods are not enough to meet the requirements for 6G networks in terms of reliability, latency, and efficiency. In fact, single-agent models of wireless problems eliminate any possibility of cooperation or coordination in the network since the agent considers all the other agents as a part of the environment. For this reason, we highlight the importance of MARL, particularly cooperative MARL, in the development of scalable and decentralized systems for 6G networks. In this context, \cite{althamary2019surveyVehicular} showcases the potential applications of MARL to build decentralized and scalable solutions for vehicle-to-everything problems. In addition, the authors in \cite{lee2020Optimization} provide an overview of the evolution of cooperative MARL with an emphasis on distributed optimization. Our work does not only consider cooperative MARL but also MBRL as enabling techniques for future 6G networks and we focus on delivering a more applied perspective of MARL to solve wireless communication problems. Table \ref{tab:SOTA} summarizes the existing surveys on DRL and 6G and highlights the key differences compared to our work. 

\subsection{Contributions and Organization of this Paper}

The main contributions of this paper can be summarized as follows:
\begin{itemize}
    \item We provide a comprehensive tutorial on single-agent DRL frameworks. Model-free RL (MFRL) is based on learning whereas MBRL is based on planning. To the best of our knowledge, this is the first initiative to present MBRL fundamentals and potentials in future 6G networks. Recent developments in the MBRL literature render these methods appealing for their sample efficiency (which is measured in terms of the minimum number of samples required to achieve a near-optimal policy) and their adaptation capabilities to changes in the environment;
    
    \item We present different MARL frameworks and summarize relevant MARL algorithms for 6G networks. In this work, we focus on the following: emergent communication where agents can learn to establish communication protocols to share information; learning cooperation details different algorithms to learn collaborative behaviors in a decentralized manner; networked agents to enable cooperation between heterogeneous agents with limited shared information;
    
    \item We also review the literature on applications of MARL in several enabling technologies for 6G networks such as MEC and control of aerial (e.g. drone-based) networks, beamforming in cell-free massive Multiple-Input Multiple-Output (MIMO) communications, spectrum management in Heterogeneous Networks (HetNets) and in THz communications, and distributed deployment and control of intelligent reflecting surface (IRS)-aided wireless systems;
    
    \item We present open research directions and challenges related to deployment of efficient, scalable, and decentralized algorithms based on RL.
\end{itemize}
\begin{table}[ht!]
    \centering
    \caption{Summary of notations and symbols}
    \label{tab:notations}
    \begin{tabularx}{0.48\textwidth}{c X}
    \specialrule{1pt}{1pt}{1pt}
    \hline
    $S$, $A$, $O$& State, action, and observation spaces \\
    $\underline{\bm{A}}, \underline{\bm{O}}$ & Joint action and observation spaces \\
    $R$, $P$ & Reward and transition functions respectively \\
    $H$ & Episode horizon or length of a trajectory  \\
    $D$ & Replay buffer \\
    $\gamma$ & Discount factor \\
    $\pi^{*}$ & Agent's optimal policy \\
    $b(s)$ & Belief state of a state $s \in S$ \\
    $\pi_{\theta}$ & Parametrized policy with parameters $\theta$ \\
    $\underline{\bm{\pi}}$ & Joint policy of multiple agents \\
    $Q^{\pi},V^{\pi}$ & $V$/$Q$-function under the policy $\pi$ \\
    $Q_{\phi}$ & Parameterized $Q$-function with parameters $\phi$ \\
    $\hat{Q},\hat{V}, \hat{\pi}$ & Approximate $V$/$Q$-function and policy\\
    $\bar{Q}$ & Target $Q$-network\\
    $J$ & Infinite-horizon discounted return \\
    $A\text{dv}$ & Advantage function \\
    \hline
    \end{tabularx}
\end{table}
\begin{table*}[ht!]
    \centering
    \caption{Summary of existing surveys on DRL and MARL for 5G and beyond wireless networks}
    \label{tab:SOTA}
    \begin{tabularx}{\textwidth}{cXcccl}
    \specialrule{1pt}{1pt}{1pt}
        \hline
        \multicolumn{1}{c}{Paper} & \multicolumn{1}{c}{Summary} & \multicolumn{3}{c}{RL techniques} & \multicolumn{1}{c}{Scope}\\
        \cline{3-5} 
        & & DRL & MBRL & MARL\\
        \hline
        \cite{luong2019Survey} & A comprehensive overview of single-agent DRL algorithms for communications and networking problems & \cmark & \xmark & \xmark & Wireless Networks \\
        \cite{qian2019surveyDRL} & Review of the recent DRL contributions in MEC, Software Defined Network (SDN) and network virtualization in 5G & \cmark & \xmark & \xmark & 5G networks\\
        \cite{lei2020AIOTDRL} & Survey on DRL applications and challenges in Autonomous IoT & \cmark & \xmark & \xmark & Autonomous IoT \\
        \cite{lee2019survey5GHetNets} & Summary of DRL applications in resource management in 5G HetNets & \cmark & \xmark & \xmark & 5G HetNets \\
        \cite{zhu2018DRLcaching} & Survey on DRL for mobile edge caching & \cmark & \xmark & \xmark & Mobile Edge Caching \\
        \cite{ali20206g} & Overview of different ML techniques and present potential applications in different network layers & \cmark & \xmark & \xmark & 6G Wireless Networks \\
        \cite{tang20196GUAV} & Review of ML and DRL applications in 6G vehicular networks & \cmark & \xmark & \xmark & 6G Vehicular networks\\
        \cite{she2020tutorial6GURLLC} &  A comprehensive tutorial on URLLC communications with a focus on the role of DL and DRL for achieving URLLC communiations in future 6G networks & \cmark & \xmark & \xmark & URLLC communications \\
        \cite{althamary2019surveyVehicular} & A survey on applications of MARL in addressing vehicular networks related issues & \cmark & \xmark & \cmark & 5G Vehicular Networks \\
        \cite{lee2020Optimization} & An overview of the evolution of cooperative MARL with an emphasis on distributed optimization & \cmark & \xmark & \cmark & Wireless Communication \\
        \cellcolor{mygreen!25}\textbf{Our work} & \cellcolor{mygreen!25}A complete and comprehensive overview of DRL methods to build decentralized solutions for B5G/6G networks & \cellcolor{mygreen!25}\cmark & \cellcolor{mygreen!25}\cmark & \cellcolor{mygreen!25}\cmark & \cellcolor{mygreen!25} B5G/6G Wireless Networks \\
        \hline
        \end{tabularx}
\end{table*}

\begin{table*}[ht!]
    \centering
    \caption{Summary of abbreviations}
    \label{tab:Abbreviations}
    \begin{tabularx}{\textwidth}{c X c X}
    \specialrule{1pt}{1pt}{1pt}
    \hline
    AI &Artificial Intelligence & MDP & Markov Decision Process \\
    A2C & Advantage  Actor-Critic  & A3C & Asynchronous Advantage Actor-Critic \\
    B & The Bellman Operator & MC & Monte Carlo \\
    CSI & Channel State Information & MEC & Mobile  Edge  Computing \\
    CTDE & Centralized Training, Decentralized Execution & MFRL & Model-Free RL \\
    Dec-POMDP & Decentralized  Partially  Observable  Markov  Processes & MG & Markov Games \\
    DQN & Deep $Q$-Networks & mMTC & Massive Machine-Type Communications \\
    DRL &  Deep Reinforcement Learning& MADDPG & Multi-Agent Deep Deterministic Policy Gradient \\
    DP & Dynamic Programming & NN & Neural Networks \\ 
    DDPG & Deep Deterministic Policy Gradient & PG & Policy Gradients \\
    DPG & Deterministic Policy Gradient & POMDP & Partially Observable Markov Decision Process \\
    D2D & Device-to-Device &POSG & Partially  Observable  Stochastic  Game \\
    eMBB & Enhanced Mobile Broadband & SG & Stochastic Games \\
    HetNets & Heterogeneous Networks & TD & Temporal Difference \\
    IL & Independent Learners & UAV & Unmanned  Aerial  Vehicles \\
    IRS & Intelligent Reflecting & URLLC & Ultra-Reliable, Low Latency Communications\\
    MAL & Multi-Agent Learning & V2V & Vehicle-to-Vehicle  \\
    MARL & Multi-Agent Reinforcement Learning & MMDP & Multi-agent Markov Decision Process\\
    MBRL & Model-Based RL & XRL & eXplainable RL \\
    
    \hline
    \end{tabularx}
\end{table*}

The rest of the paper is organized as follows. In Section~\ref{sec:background}, we introduce the mathematical background for both single-agent RL and MARL. Standard algorithms for single-agent RL are reviewed in Section~\ref{sec:MFRL}. In Section~\ref{sec:MBRL}, we introduce MBRL and detail potential applications for 6G systems. Section \ref{sec:MARL} first summarizes the different challenges of MARL and afterward dwells on the cooperative MARL algorithms according to the type of cooperation they address. Section \ref{Sec:App} is dedicated to recent contributions of the mentioned algorithms in several wireless communication problems, followed by a conclusion and future research directions outlined in Section \ref{Sec:Conclusion}. A summary of key notations and symbols is given in Table \ref{tab:notations}.

\section{Background}
\label{sec:background}
The objective of this section is to present the mathematical background and preliminaries for both single agent and multi-agent RL.
\subsection{Single-Agent Reinforcement Learning}
\subsubsection{\textbf{Markov Decision Process}}
\label{sec:MDP}
In RL, a learning agent interacts with an environment to solve a sequential decision-making problem. \emph{Fully observable} environments are modeled as MDPs defined as a tuple $(S, A, P ,R, \gamma)$. $S$ and $A$ define the state and the action spaces respectively; $P := S \times A \mapsto[0,1]$ denoted the probability of transiting from a state $s$ to a state $s^{\prime}$ after executing an action $a$; $R: S \times A \times S \mapsto  \mathbb{R}$ is the reward function that defines the agent's immediate reward for executing an action $a$ at a state $s$ and resulting in the transition to $s^{\prime}$; and $\gamma \in [0,1]$ is a discount factor that trades-off the immediate and upcoming rewards. 
The full observability assumption of MDPs enables the agent to access the exact state of the system $s$ at every time step $t$. Given the state $s$, the agent will \emph{decide} to take an action $a$ transiting the system to a new state $s^{\prime}$ sampled from the probability distribution $P(.|s, a)$. The agent will be rewarded with an immediate compensation $R(s, a, s^{\prime})$. Thus, the agent's expected \textbf{return} is expressed as $\mathbb{E} \Big[\sum_{t=0}^{\infty} \gamma^t R(s, a, s^{\prime}) | a \sim \pi(.|s), s_0 \big] $. This is referred to as \emph{infinite-horizon discounted} return. Another popular formulation is \emph{undiscounted finite-horizon} return $\mathbb{E} \Big[\sum_{t=0}^{H} R(s, a, s^{\prime}) | a \sim \pi(.|s), s_0 \big] $ where the return is compute over a finite horizon $H$. This is common in {\em episodic tasks} (i.e. tasks that have an end). Note that the finite-horizon setting can be viewed as infinite-horizon case by augmenting the state space with an absorbing state transiting continuously to itself with zero rewards.

The agent aims to find an optimal policy $\pi^*$, a mapping from the environment states to actions, that maximizes the expected return. A policy or a strategy describes the agent's behaviour at every time step $t$. A \emph{deterministic} policy returns actions to be taken in each perceived state. On the other hand, a \emph{stochastic} policy outputs a distribution over actions. Under a given policy $\pi$, we can define a \emph{value function} or a \emph{$Q$-function} which measures the expected accumulated rewards staring from any given state $s_t$ or any pair $(s_t, a_t)$ and following the policy $\pi$, as shown below:
\begin{equation*}
\resizebox{.92\hsize}{!}{$V^{\pi}(s) = \mathbb{E} \Bigg[\sum_{t=0}^{\infty} \gamma^t R(s_t, a_t, s_{t+1}) | a_t \sim \pi(.|s_t), s_0=s \Bigg]$}
\end{equation*}
\begin{equation*}
\resizebox{.92\hsize}{!}{$Q^{\pi}(s,a) = \mathbb{E} \Bigg[\sum_{t=0}^{\infty} \gamma^t R(s_t, a_t, s_{t+1}) | a_t \sim \pi(.|s_t), s_0=s, a_0=a \Bigg]$}.
\end{equation*}
Using Dynamic Programming (DP) methods such as \emph{Value Iteration} or \emph{Policy Iteration} \cite{sutton2018reinforcement} to solve an MDP mandates that the dynamics of the environment ($P$ and $R$) are known which is often not possible. This motivates the model-free RL approaches that find the optimal policy without knowing the world's dynamics. MBRL methods learn a model of the environment by estimating the transition function and/or the reward function and use the approximate model to learn or improve a policy. Model-free RL algorithms are discussed in detail in Section~\ref{sec:MFRL} and MBRL is further investigated in Section~\ref{sec:MBRL}.

\vspace{0.2cm}
\noindent
\textbf{Example}:
Several wireless problems have been formulated as MDPs. As an example, \cite{ahmed2019MDPPowAllo} presented downlink power allocation problem in a multi-cell environment as an MDP. The agent is an ensemble of $K$ base stations (or a controller for $K$ base stations). The state space consists of the users' channel quality and their localization with respect to a given base station. The agent selects the power levels for $K$ base stations to maximize the entire network throughput. 

\subsubsection{\textbf{Partially Observable Markov Decision Process}}
\label{sec:POMDP}
In the previous section, it was assumed that the agent has access to the \emph{full} state information. However, this assumption is violated in most real-world applications. For example, IoT devices collect information about their environments using sensors. The sensor measurements are noisy and limited, hence the agent will only have \emph{partial} information about the world. Several problems such as \emph{perceptual aliasing} prevent the agent from knowing the full state information using the sensors' observations. 
In this context, \emph{Partially Observable Markov Decision Processes} (POMDP) generalize the MDP framework to take into account the uncertainty about the state information. POMDP is described by a 7-tuple $(S, A, P, R, \gamma, O, Z)$ where the first five elements are the same as defined in \S\ref{sec:MDP}; $O$ is the observations space and $Z: S \times A \times O \mapsto [0,1]$ denotes the probability distribution over observations given a state $s \in S$ and an action $a \in A$.
To solve a POMDP, we distinguish two main approaches. The first one is \emph{history-based} methods where the agent maintains an observation history $H_t = \{o_1, \dots, o_{t-1}\}$ or an action-observation history $H_t = \{(o_0, a_0), \dots, (o_{t-1}, a_{t-1})\}$. The history is used to learn a policy $\pi(.|H_t)$ or a $Q$-function $Q(H_t, a_t)$. As an analogy with MDPs, the agent state becomes the history $H_t$. As a result, this method has a large state space which can be alleviated by using a truncated version of the history (i.e. $k$-order Markov model). However, limited histories have also caveats since long histories need more computational power and short histories suffer from possible information loss. It is not straightforward how the value of $k$ is chosen. Another way to avoid the increasing dimension of the history with time is by defining the notion of \emph{belief state} $b_t(s) = p(s|H_t), \forall s\in S$ as a distribution over states. Thus, the history $H_t$ is indirectly used to estimate the probability of being at a state $s$. Therefore, a $Q$-function $Q(b(s), a)$ or a policy $\pi(.|b(s))$ (see Table~I) can be learned using the belief states instead of the history. If the POMDP is known, the belief states are updated using Bayes' rule. Otherwise, a Bayesian approach can be considered. Another approach to solve a POMDP is \emph{Predictive State Representation} \cite{littman2002PSR}. The main idea consists in predicting what will happen in the future instead of relying on past actions and observations. 

\vspace{0.2cm}
\noindent
\textbf{Example}:
POMDP formulation was applied to solve different wireless problems characterized by partial access to the environment state. For example, \cite{xie2020ExPOMDP} proposed a POMDP representation of dynamic task offloading in IoT fog systems. The authors assume that the IoT devices have imperfect Channel State Information (CSI). In this scenario, the agent is the IoT device, and based on the estimated CSI and the queue state, it decides the tasks to be executed locally or offloaded. POMDPs are widely used in wireless sensor networks.

\subsection{Multi-Agent Reinforcement Learning}
MARL tackles sequential decision-making problems involving a set of agents. Hence, the system dynamics is influenced by the joint action of all the agents. More intuitively, the reward received by an agent is no longer a function of its own actions but a function of all the agents' actions. Therefore, to maximize the long-term reward, an agent should take into consideration the policies of the other agents. In what follows, we will present mathematical backgrounds for MARL. Please refer to Section~\ref{Sec:App} for examples on how to formulate wireless communication problems using the discussed mathematical frameworks below. 

\vspace{0.2cm}
\subsubsection{\textbf{Markov/Stochastic Games}}
\label{MG}
MGs or SGs \cite{shapley1953stochastic} extend the MDP formalism to the multi-agent setting to take into account the relation between agents \cite{littman1994markov}. Let $N > 1$ be the number of agents, $S$ is the state space and $A_i$ denotes the action space of the $i$'s agent. The joint action space of all agents is given by $\underline{\bm{A}} := A_1\times\dots\times A_N$. From now on, we will use bold and underlined characters to differentiate between joint and individual functions.

At a state $s$, each agent $i$ selects an action $a_i$ and the joint action $\underline{\mathbf{a}} = [a_i]_{i \in N}$ will be executed in the environment. The transition from the state $s$ to the new state $s^{\prime}$ is governed by the transition probability function $P : S \times \underline{\bm{A}} \times S \mapsto [0,1]$. Each agent $i$ will receive an immediate reward $r_i$ defined by the reward function $R_i : S \times \underline{\bm{A}} \times S \mapsto \mathbb{R}$. Therefore, the MG is formally defined by the tuple $(N, S, (A_i)_{i\in N} , P , (R_i )_{i\in N}, \gamma)$ where $\gamma$ is a discount factor. Note that the transition and reward functions in MG are dependent on the joint action space $\underline{\bm{A}}$. 
Each agent $i$ seeks to find the optimal policy $\pi_i^* : S\mapsto A_i$ that maximizes its long-term return. The joint policy $\underline{\bm{\pi}}$ of all agents is defined as $\underline{\bm{\pi}}(\underline{\mathbf{a}}|s) = \prod_{i\in N} \pi_i(a^i|s)$. Hence, the value-function of an agent $i$ is defined as follows:
\begin{equation*}
\resizebox{.96\hsize}{!}{$V_i^{\underline{\bm{\pi}}}(s) =  \mathbb{E}_{\underline{\bm{\pi}}}\Bigg[ \sum_{t=0}^{\infty} \gamma^t R_i(s_t, \underline{\bm{a}}_t, s_{t+1}) | \underline{\bm{a}}_t \sim \underline{\bm{\pi}}(s_t),s_{0}=s \Bigg]$}.
\end{equation*}
Consequently, the optimal policy of the agent $i$ is a function of its opponents' policies $\pi_{-i}$. 

The complexity of MARL systems arises from this property because the other agents' policies are non-stationary and change during learning. See Section \ref{sec:MARLchallenges} for a detailed discussion of MARL challenges.
As mentioned before, we distinguish three solution concepts for MGs: \emph{fully-cooperative}, \emph{fully competitive}, and \emph{mixed}. In fully cooperative settings, all the agents have the same reward function $R_i=R$ and hence the same value or state-action function. Fully-cooperative MGs are also referred to as \emph{Multi-agent MDP} (MMDP). This simplifies the problem since standard single-agent RL algorithms can be applied if all the agents are coordinated using a central unit. On the other hand, fully competitive MGs ($\sum_i Ri = 0$) and general-sum MGs ($\sum_i Ri \in \mathbb{R}$) are addressed by searching for a NE. We will focus on the subsequent sections on extensions of MG for cooperative problems.

\vspace{0.2cm}
\noindent
\textbf{Example}: MGs are the most straightforward generalization of single-agent wireless problems to the multi-agent scenarios. As an example, the problem of field coverage by a team of UAVs is modeled as MG in \cite{UAVPham2018}.

\begin{figure}[ht]
\begin{subfigure}{.241\textwidth}
  \centering
  \includegraphics[scale=0.5]{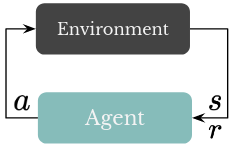}
  \caption{MDP}
  \label{fig:MDP}
\end{subfigure}%
\begin{subfigure}{.248\textwidth}
  \centering
  \includegraphics[scale=0.5]{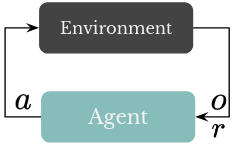}
  \caption{POMDP}
  \label{fig:POMDP}
\end{subfigure}
\begin{subfigure}{.242\textwidth}
  \centering
  \includegraphics[scale=0.5]{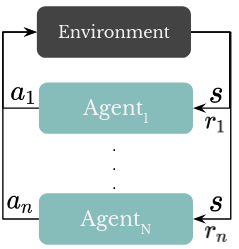}
  \caption{Markov Games}
  \label{fig:MG}
\end{subfigure}
\begin{subfigure}{.24\textwidth}
  \centering
  \includegraphics[scale=0.5]{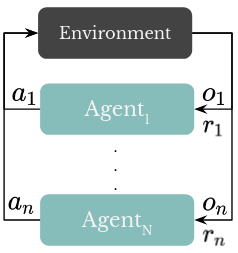}
  \caption{Dec-POMDP}
  \label{fig:decPOMDP}
\end{subfigure}
\caption{Mathematical frameworks described in Section \ref{sec:background}.}
\label{fig:rlFrameworks}
\end{figure}


\vspace{0.2cm}
\subsubsection{\textbf{Dec-POMDP}}
The intertwinement of agents in the multi-agent setting adds more complexity to finding optimal policies for each agent. In fact, each agent needs full information about the other agents' actions to maximize its long-term rewards. Consequently, the uncertainty about the other opponents in addition to the state uncertainty call for an extension of the MG framework to model cooperative agents under partial observable environments. In this context, \emph{Decentralized Partially Observable Markov Decision Process (Dec-POMDP)} \cite{oliehoek2016DECPOMDP} is the adopted mathematical framework to study the cooperative sequential decision-making problems under uncertainty. This is a direct generalization of POMDPs to the multi-agent settings. A Dec-POMDP is described as $(N, S, (A_i)_{i\in N}, P, R, \gamma, (O_i)_{i\in N}, Z)$ where the six first elements are same as defined in \S\ref{MG}; $R$ is a global reward function shared by all the agents; $O_i$ is the observation space of the $i$'s agent with $\underline{\bm{O}}:= O_1 \times \dots \times O_N$ is the joint observation space and $Z: S \times \underline{\bm{A}} \times \underline{\bm{O}} \mapsto [0,1]$ is the observation function which provides the probability $P(\underline{\bm{o}}|\underline{\bm{a}}, s^{\prime})$ of the agents observing $\underline{\bm{o}}=[o_1 \times \dots \times o_N]$ after executing a joint action $\underline{\bm{a}} \in \underline{\bm{A}}$ and transiting to a new state $s^{\prime} \in S$. A Dec-POMDP is a specific case of the \emph{Partially Observable Stochastic Games} (POSG) \cite{hansen2004posg} defined as a tuple $(N, S, (A_i)_{i\in N}, P, (R_i)_{i\in N}, \gamma, (O_i)_{i\in N}, Z)$ where all the elements are the same as in Dec-POMDP expect the reward function $R_i$ which becomes individual for each agent. POSG enables the modeling of self-interest agents whereas Dec-POMDP exclusively models cooperative agents in partial observable environments. At a state $s$, each agent $i$ receives its own observation $o^{i}$ without knowing the other agents' observations. Thus, each agent $i$ chooses an action $a^{i}$ yielding a joint action to be executed in the environment. Based on a common immediate reward, each agent strives to find a \emph{local} policy $\pi_i : O_i \mapsto A_i$ that maximizes its team long-term reward. Thus, the \emph{joint} policy is given by $\underline{\bm{\pi}}=[\pi_1, \dots, \pi_N]$. The policy $\pi_i$ is called \emph{local} because each agent acts according to its own local observations without communicating or sharing information with the other agents.

\vspace{0.2cm}
\noindent
\textbf{Example}: Multi-agent task offloading \cite{MECMARL1} and multi-agent cooperative edge caching \cite{CoopCachingZhong2019} are wireless problems which can be modeled as Dec-POMDP problems. 
\vspace{0.2cm}
\subsubsection{\textbf{Networked Markov Games}}
\begin{figure*}[ht]
     \centering
     \begin{subfigure}[b]{0.32\textwidth}
         \includegraphics[scale=0.5]{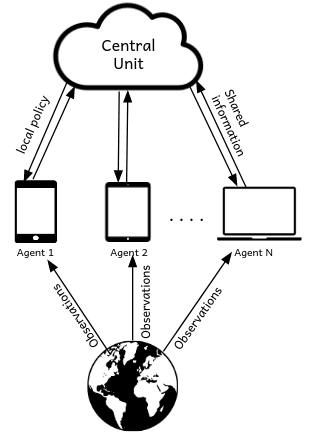}
         \caption{CTDE}
         \label{fig:centralized}
     \end{subfigure}
     \hfill
     \begin{subfigure}[b]{0.32\textwidth}
         \centering
         \includegraphics[scale=0.5]{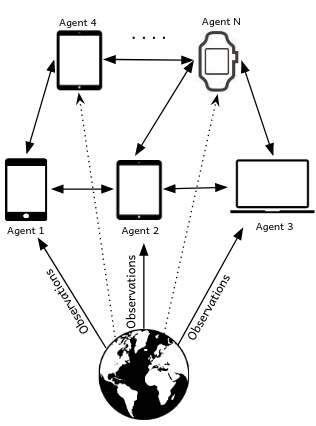}
         \caption{Networked Agents}
         \label{fig:DecNet}
     \end{subfigure}
     \hfill
     \begin{subfigure}[b]{0.32\textwidth}
         \centering
         \includegraphics[scale=0.5]{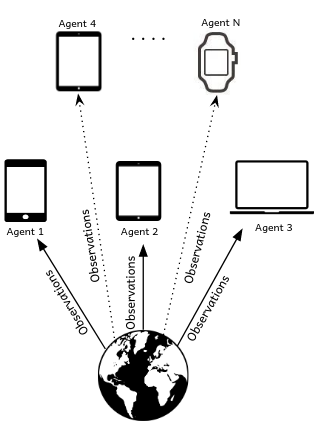}
         \caption{Fully Decentralized}
         \label{fig:fullyDecentralized}
     \end{subfigure}
        \caption{Three representative learning frameworks in MARL. Specifically, in (a), the \emph{centralized training and decentralized execution} scheme is characterized by a central unit collecting information from the agents (i.e observations, joint actions) to train the agents' local policies. During execution, the agents only need their local information to act in the system. This scheme has a considerable communication cost which will increase with the number of users. In both (b) and (c), we present two decentralized learning structures where no central unit is needed. Figure (b) illustrates the \emph{decentralized over Networked Agents} scheme where agents are able to communicate with their neighbors over a possibly time-varying network. This allows the local information exchange. In (c), the \emph{fully decentralized} or \emph{independent learners} scheme is represented. In this learning framework, no explicit information exchange takes place between the agents.}
        \label{fig:Marl_training}
\end{figure*}

Cooperative MGs or Dec-POMDPs are only suitable for \emph{homogeneous} cooperative agents since they share the same reward signal $R_1=\dots=R_N= R$. However, most of real-world applications involve \emph{heterogeneous} agents with distinct preferences and goals. In addition, sharing the same reward function requires global information from all the agents to estimate a global value or a state-action functions which complicates the decentralization of such models. To overcome these shortcomings, \emph{Networked MG} generalizes the MG framework to model cooperative agents with different reward functions by leveraging shared information through a communication network (see Figure \ref{fig:Marl_training}.b). Formally, Networked MGs are described as a tuple $(N, S, (A_i)_{i\in N}, P, (R_i )_{i\in N}, (G_t)_{t \geq 0})$ where the first five elements are same as defined in \S\ref{MG} and $G_t = (N, \mathcal{E}_t)$ is a time-varying communication network linking $N$ nodes with a set of edges $\mathcal{E}_t$ at time $t$. An edge $(i,j) \in \mathcal{E}_t, \forall i,j$ means that both agents $i$ and $j$ can communicate and share information mutually at time $t$. Henceforth, agents know their local and neighboring information and seek to learn an optimal joint policy by maximizing the team-average reward $\bar{R}(s, a, s^{\prime}) = \frac{1}{N-1} \sum_{i\in N} R_i(s, a, s^{\prime})$ for any $(s, a, s^{\prime}) \in S \times \underline{\bm{A}} \times S$. To summarize, the advantages of Networked MGs compared to classical MGs are: (i) the possibility to model heterogeneous agents with different reward functions; (ii) the reduction of the coordination cost by considering neighbor-to-neighbor communication which facilitates the design of decentralized MARL algorithms; (iii) the privacy preserving property since agents are not mandated to share their reward functions.

\vspace{0.2cm}
\noindent
\textbf{Example}:
Networked MDP can be applied in multiple wireless scenarios where agents are linked with a communication graph. For example, base stations in cell-free networks can collaborate to compute optimal beamforming while minimizing interference~\cite{Yesser2020Beamforming}. The communication graph will enable the base stations to share information with their neighbors. Thus, better collaboration is possible.


\section{Single Agent Model-Free RL Algorithms}
\label{sec:MFRL}

\subsection{Preliminaries}
We start by defining useful notions and concepts for the understanding of the algorithms discussed below. 

MFRL methods can be categorized into two classes depending on the agent's learned objective. In \emph{value-based} methods, an approximate value function $\hat{V}$ is learned and the agent's policy $\pi$ is obtained by acting \emph{greedily} with respect to $\hat{V}$. Thus, state-values are essential for action selection. \emph{Policy evaluation} methods seek to learn an estimate of the value function $\hat{V} = V^\pi $ for a given policy $\pi$. Alternatively, \emph{policy-based} methods aim to directly learn a \emph{parameterized} policy without resorting to a value function. A well-known variant of policy-based methods learns an approximation of the value function but the action selection is still independent of the value estimates. These are the \emph{actor-critic} methods where approximations to both the policy and the value function are learned. The actor refers to the policy and the critic is the approximate value function. Henceforth, we will denote by $\theta$ the policy's parameters and $\pi_\theta$ is the policy parametrized by $\theta$. In actor-critic methods, $V^{\pi_\theta}_\phi$ denotes the approximate value function under the policy $\pi_\theta$ where $\phi$ is a learnable parameter vector. 

We distinguish between two main learning principles: (i) \emph{Monte Carlo} (MC) and (ii) DP methods. The former methods utilize experience to approximate value functions and policies. In contrast, DP methods are known for solving the \emph{Bellman Optimality} equations. More details will be provided in the following sections. \emph{Temporal  Difference} (TD) is a famous combination of these two learning frameworks. Therefore, an important question arises when MC and TD methods are adopted: how actions are selected and samples are generated for learning?
This leads to the two key approaches for learning from experience, namely, \emph{off-policy} and \emph{on-policy} methods. Recall that the agent interacts with its environment by executing actions and after improve or evaluate its policy using the collected data. Therefore, we can distinguish between two distinct processes: the policy used for data collection and the policy being improved or evaluated. The former is called \emph{behavior policy} and the latter is referred to as \emph{target policy} or \emph{control policy}. In off-policy methods, the behavior policy is different from the target policy. However, in on-policy methods, the behavior and the target policies are the same, meaning that the policy used to collect the data samples is the same as the one being evaluated or improved. Thus, the notation $V^{\pi_\theta}_\phi$ means that the value function $V$ is learned using samples from the policy ${\pi_\theta}$. If ${\pi_\theta}$ is the same as the policy the agent is learning, this is an on-policy algorithm. The advantage of an off-policy setting is the possibility to use a more exploratory behavior policy to continue to visit all the possible actions. This is why off-policy methods encourage \emph{exploration}. Exploration-exploitation trade-off is a well-known challenge in RL: The agent can exploit the knowledge from its past experiences to choose actions with the highest expected rewards but it also needs to explore other actions to improve its action selection. 

Another key distinction between RL learning frameworks is the use of \emph{Bootstrapping}. The general idea of bootstrapping is estimated values of states are updated based on estimates of the values of the next states. DP and TD methods use bootstrapping whereas MC algorithms rely on actual complete episodic returns.

Furthermore, another dimension to consider while designing an RL algorithm is how to represent the approximate value function. In \emph{tabular} setting, a table of state-values is maintained and updated for every visited state $s$. \emph{Function approximators} have enabled the recent revolution in RL thanks to its generalization power with high-dimensional state data. For example, DNNs are famous non-linear function approximators used to compute value functions or policies. 

As mentioned above, the source of the training data is crucial for learning. On one hand, \emph{batch/offline RL} considers the agent is provided with a dataset of interactions and learns a policy using the given dataset without interacting with the environment. On the other hand, the agent can collect data by querying the real environment or a simulator. This is referred to as \emph{online RL}.

\begin{figure*}[ht!]
        \centering
         \includegraphics[scale=0.5]{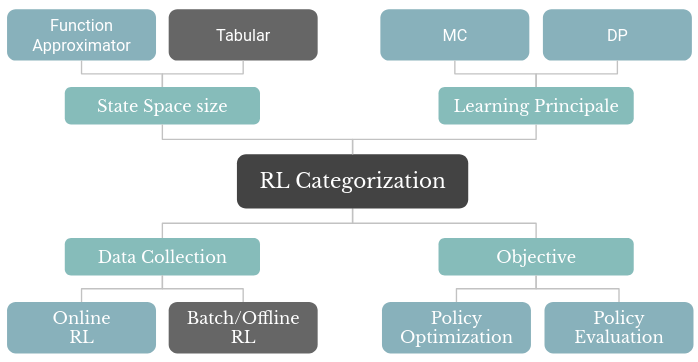}
         \caption{Categorization of different RL settings. The classes colored in blue are covered in Section \ref{sec:background}.}
         \label{fig:RLcatgs}
\end{figure*}

Figure \ref{fig:RLcatgs} summarizes the different categorization of RL methods. In this tutorial, we will focus on both online policy-based and value-based methods with DNNs as function approximators. Table \ref{tab:MFRL} provides a comparative overview of the discussed algorithms below. This review of MFRL methods is not exhaustive since several resources with in-depth descriptions are already available (i.e. in \cite{sutton2018reinforcement}, \cite{luong2019Survey}).


\subsection{Policy-Based Algorithms}
Policy-based methods directly search for the optimal policy by maximizing the agent's expected long-term reward $J$ as in  (\ref{eq:PG1}). The policy is parameterized by a function approximator $\pi_{\theta}(a|s)$, typically a DNN with learnable weights $\theta$. The \emph{Policy Gradient} (PG) methods, introduced in \cite{sutton2000PG}, learn the optimal parameters $\theta^*$ by performing gradient ascent on the objective $J$. Using the PG theorem \cite{sutton2000PG}, the policy gradients are expressed as in (\ref{eq:PG3}) and estimated using samples or trajectories collected under the current policy. This is why PG methods are on-policy methods. For each gradient update, the agent needs to interact with the environment and collect trajectories. Samples collected at iteration $k$ cannot be reused for the next policy update. This \textbf{sample inefficiency} represents one of the major drawbacks of PG methods.

\begin{align}
    J(\theta) &= \mathbb{E}_{\pi_\theta}\Bigg[\sum_{t=0}^{\infty} \gamma^t R(s_t, a_t)\Bigg]. \label{eq:PG1}\\
    \nabla_{\theta} J(\theta) &= \mathbb{E}_{\pi_\theta} \left[ 
    \sum_{t=0}^{T} \nabla \log \pi_{\theta}(a_t|s_t) Q^{\pi_\theta}(s_t,a_t) \right] \label{eq:PG3} 
\end{align}
In (\ref{eq:PG3}), $Q^{\pi_\theta}$ is not known and needs to be estimated. Several approaches are possible. The well-known \emph{REINFORCE} algorithm \cite{williams1992simple} uses the \emph{rewards-to-go} defined as $\sum_{k=t}^T R(s_k, a_k)$. The major caveat of the \emph{REINFORCE} algorithm is that it is well-defined for \emph{episodic} problems only since the rewards-to-go are computed at the end of an episode. Furthermore, REINFORCE algorithm suffers from high variance. In (\ref{eq:PG3}), action likelihoods are multiplied by their expected return thus PG algorithm shifts the action distribution such that good actions are more likely than bad ones. Consequently, small variations in the returns can lead to a completely different policy. This motivates actor-critic methods where an approximation of $Q^{\pi_\theta}$ is learned. Note that it also possible to estimate the value function $V^{\pi_\theta}$ or the advantage function $A\text{dv}^{\pi_\theta} = Q^{\pi_\theta} - V^{\pi_\theta}$. Learning a critic reduces the variance of gradient estimates since different samples are used whereas in the rewards-to-go only one sample trajectory is considered. However, actor-critic methods introduce bias since the $Q^{\pi_\theta}$ estimate can be biased as well. In this context, \cite{schulman2015high} proposed \emph{Generalized Advantage Estimation} based on the idea of n-step returns to reduce the bias. In what follows, we will examine the most common policy gradient algorithms.

Asynchronous Advantage Actor-Critic (A3C) \cite{mnih2016A3C} proposes a parallel implementation of the actor-critic algorithm. In the original version of A3C, a global NN outputs the action probabilities and an estimate of the agent's advantage function. Thus, the actor and the critic share the network layers. Several workers are instantiated with local copies of the global network parameters and the environment. These workers are created as CPU threads in the same machine. In parallel, each worker interacts with its local environment and collects experiences to estimate the gradients with respect to the network parameters. Afterward, the worker propagates its gradients and updates the parameters of the global network. Therefore, the global model is constantly updated by the workers. This learning scheme enables the collection of more diverse experiences since each worker interacts with their local copy of the environment independently. The drawback of the asynchronous training scheme is that some workers will be using old versions of the global network. In this context, Advantage Actor-Critic (A2C) adopts a synchronous and deterministic implementation where all the workers' gradients are aggregated and averaged to update the global network.

As mentioned before, PG algorithms suffer from sample inefficiency since only one gradient update is performed per a batch of collected data. This motivates the goal to use the data more efficiently. Besides, it is hard to pick the learning rate since it affects the training performance and can dramatically alter the visitation distribution. Intuitively, a high learning rate can result in a bad policy update which means that the next batch of data is collected using a bad policy. Recovering from a bad policy update is not guaranteed. This motivate Trust Region Policy Optimization (TRPO) \cite{schulman2015TRPO} where the original optimization problem in (\ref{eq:PG1}) is solved under the constraint of ensuring the new updated policy is close to the old one. To do so, the constraint is defined in terms of Kullback–Leibler divergence ($D_\text{KL}$) which measures the difference between two probability distributions. More formally, let $\theta_k$ be the policy parameters at  iteration $k$. We would like to find the new parameters $\theta_{k+1}$ such that \begin{align*}
    \theta_{k+1} & = \argmax_{\theta}L(\theta) \\
    &= \argmax_{\theta} \mathbb{E}_{(s,a) \sim \pi_{\theta_k}} \bigg[ \frac{\pi_\theta(a|s)}{\pi_{\theta_k}(a|s)} A\text{vd}^{\pi_{\theta_k}}(s,a)\bigg]  \\
    & \text{s.t.} \quad D_\text{KL}(\theta || \theta_k) \leq \delta,
\end{align*}
where $\delta$ is the trust region radius. Let $F$ be the Fisher-information matrix. With a first-order approximation of the objective ($L(\theta) \approx \nabla_\theta L^{T}(\theta)(\theta - \theta_k)$) and a second-order Taylor expansion of the constraint ($D_\text{KL}(\theta || \theta_k) \approx \frac{1}{2} (\theta - \theta_k)^TF(\theta - \theta_k)$), the update rule is given by $\theta_{k+1} = \theta_k + \sqrt{\frac{2\delta}{\nabla_\theta L^{T}(\theta) F^{-1}\nabla_\theta L(\theta)}}F^{-1}\nabla_\theta L(\theta)$. The term $F^{-1}\nabla_\theta L(\theta)$ is called the \emph{natural gradients}. Consequently, evaluating the natural gradients necessitates inverting the matrix $F$ which is expensive. To overcome this issue, TRPO implements the conjugate gradient algorithm to solve the system $Fx = \nabla_\theta L(\theta)$ which involves evaluating $Fx$ instead. Finally, the matrix-vector product $Fx$ is computed as $\nabla_\theta(\nabla_\theta D_\text{KL}(\theta || \theta_k)^Tx)$ which is easy to evaluate using any auto-differentiation library like Tensorflow. 
In a similar vein, Proximal Policy Optimization (PPO) \cite{schulman2017PPO} algorithm solves the same optimization problem as TRPO but proposes a simpler implementation by introducing a new loss function:
\begin{align*}
    L_\text{PPO}(\theta) = \min_\theta \bigg(& \frac{\pi_\theta(a|s)}{\pi_{\theta_k}(a|s)} A\text{vd}^{\pi_{\theta_k}}(s,a), \\
    & \text{clip}(\frac{\pi_\theta(a|s)}{\pi_{\theta_k}(a|s)}, 1-\delta, 1+\delta)A\text{vd}^{\pi_{\theta_k}}(s,a) \bigg),
\end{align*}
where ``clip" is a function used to keep the values of the ratio $\frac{\pi_\theta(a|s)}{\pi_{\theta_k}(a|s)}$ between $1-\delta$ and $1+\delta$ to penalize the new policy if gets far from the old policy.

\vspace{0.2cm}
\noindent
\textbf{Example}: The advantage of policy-based algorithms is they are applicable to discrete and continuous action spaces. For example, \cite{valadarsky2017TRPORouting} applies the TRPO algorithm to find optimal routing strategies in a network.
\subsection{Value-Based Algorithms}
\label{sec:QL}
As explained above, value-based algorithms focus on estimating the agent's value function. Thus, the policy is computed implicitly or \emph{greedily} with respect to the approximate value function. 

The MC method approximates the value of a state $s$ by averaging the rewards obtained after visiting the state $s$ until the end of an episode. Consequently, MC methods are defined only for episodic tasks. In DP, the optimal value $V^*$ and state-action $Q^*$ function are computed by solving the \emph{Bellman Optimality} equations (\ref{eq:vstar}-\ref{eq:qstar}) and thus the optimal policy is obtained greedily with respect to the Q-values (\ref{eq:pstar}):
\begin{align}
    V^*(s) &= \max_a \big[R(s,a) + \gamma \mathbb{E}_{s^\prime}V^*(s^\prime)\big] \label{eq:vstar}\\
    Q^*(s) &= R(s,a) + \gamma \mathbb{E}_{s^\prime}\big[\max_{a^\prime} Q^*(s^\prime, a^\prime)\big] \label{eq:qstar}\\
    \pi^{*}(s) &= \argmax_a Q^*(s,a). \label{eq:pstar}
\end{align}

Let $B:\mathbb{R}^{|S\times A|} \mapsto \mathbb{R}^{|S\times A|}$ denote the \emph{Bellman optimality} operator such that $[BQ](s,a) = R(s,a) + \gamma \sum_{s^\prime} P(s^\prime|s,a) \max_{a^\prime} Q^*(s^\prime, a^\prime)$. Therefore, equation (\ref{eq:qstar}) can be written in more compact way $Q^*=BQ^*$. As a result, $Q^*$ is the called the \emph{fixed point} of the Bellman optimality operator and the methods solving for this fixed point can be called fixed point methods. The value iteration algorithm approximates $Q^*$ by iteratively applying the Bellman optimality operator $\hat{Q}_k = B\hat{Q}_{k-1}$. This algorithm is guaranteed to converge to $Q^*$ since $B$ is a \emph{contraction} and $Q^*$ always exists and is unique. Besides, value iteration relies on bootstrapping to estimate the value of next states. However, to evaluate the Bellman operator, the transition function is needed. This is the major drawback of DP methods that assume the environment dynamics are known. To overcome this issue, TD methods combine the main ideas of MC and DP. They use experience as in MC and bootstrapping as in DP. The update rule of TD algorithm is as follows:
\begin{equation}
\label{eq:tdupdate}
\resizebox{.91\hsize}{!}{$\hat{Q}(s,a) = (1-\alpha)\hat{Q}(s,a) + \alpha \big[R(s,a) + \gamma \max_{a^\prime} \hat{Q}(s^\prime,a^\prime)\big]$,}
\end{equation}
where $\hat{V}$ and $\hat{Q}$ are the approximate value and state-action functions and $\alpha$ is a learning rate. 

TD methods can be on-policy or off-policy. Let $\hat{\pi}$ the policy derived from $\hat{Q}$ (i.e. $\epsilon$-greedy). For on-policy TD, the samples used to estimate $\hat{Q}$ are generated using the current policy $\hat{\pi}$ continuously updated greedily with respect to $\hat{Q}$. \emph{SARSA} is a well-known on-policy TD algorithm where the agent collects experiences in the form $\{(s,a,r,s^\prime,a^\prime)\}$. Since the action in the next state is known, the $\max$ operator in the RHS of the TD update (\ref{eq:tdupdate}) is removed. 

$Q$-learning algorithm \cite{watkins1992q} revolutionized the RL world allowing the development of an \emph{off-policy} TD algorithm. Any policy $\tilde{\pi} \neq \hat{\pi}$ can be used to generate experiences. When $\hat{Q}$ is represented with a function approximator with parameters $\phi$, $Q$-learning algorithm minimizes the \emph{Bellman error} (\ref{eq:5}) and updates the $Q$-function parameters as in (\ref{eq:q6}). The Bellman error is not a contraction anymore thus the convergence guarantees discussed earlier are not valid anymore. Equation (\ref{eq:qtarget}) defines the \emph{targets}. Note the update rule in (\ref{eq:q6}) does not consider the full gradient of the Bellman error since it ignores the gradients of the targets with respect to the parameters $\phi$. This is why, this learning algorithm is also called \emph{semi-gradient} method~\cite{sutton2018reinforcement}. 
\begin{align} 
    \phi^* &=  \argmin_\phi \frac{1}{2} \sum_{(s,a,r,s^{\prime})} ||\hat{Q}_{\phi}(s, a) - y||^2. \label{eq:5} \\ 
    y &=  R(s,a) + \gamma \max_{a^{\prime}} \hat{Q}_{\phi}(s^{\prime},a^{\prime}). \label{eq:qtarget}\\
    \phi & = \phi - \alpha  \sum_{(s,a,r,s^{\prime})}  \nabla_\phi Q_\phi(s,a)\bigg(Q_\phi(s,a) - y\bigg). \label{eq:q6}
\end{align} 
Deep versions of the $Q$-learning methods  such as Deep $Q$-Networks (DQN) \cite{mnih2013DQN} have been developed. In particular, the $Q$-function is parameterized using a DNN with weights $\phi$. DQN and its variants are the most popular online $Q$-learning algorithms that have shown impressive results in several communication applications. To stabilize the learning using DNN, DQN introduces  techniques such as an \emph{experience replay buffer} $D=\{(s,a,r,s^\prime)\}$ to avoid correlated samples and get a better gradient estimation and a target network $\Bar{Q}$, whose parameters $\phi^{'}$ are periodically updated with the most recent $\phi$, making the targets (\ref{eq:qtarget}) stationary and do not depend on the learned parameters. The target network is updated periodically as follows:
\begin{align} 
    y_\text{DQN} &=  R(s,a) + \gamma \max_{a^{\prime}} \bar{Q}_{\phi^\prime}(s^{\prime},a^{\prime}). \label{eq:dqntarget} \\
    y_{\text{DDQN}} &=  r(s,a) + \gamma \hat{Q}_2(s^\prime, \argmax_{a^{\prime}}\hat{Q}_1(s^{\prime},a^{\prime})). \label{eq:ddqntarget}
\end{align} 
Although $Q$-learning algorithms are sample efficient, they lack convergence guarantees for non-linear function approximators and also suffer from the maximization bias which results in an unstable learning process. The $\max$ operator in equation (\ref{eq:qtarget}) makes the algorithm overestimate the true $Q$-values and is also problematic for continuous action spaces. To overcome the maximization bias, double learning technique uses two networks $\hat{Q}^1_{\phi_1}$ and $\hat{Q}^2_{\phi_2}$ with different parameters and decouples the action selection from the action evaluation as shown in (\ref{eq:ddqntarget}). Other variants of DQN have been suggested to guarantee a better convergence. \emph{Prioritized experience replay} is proposed in \cite{schaul2015prioritized} to ensure the sampling of rare and task-related experiences is more frequent than the redundant ones. \emph{Dueling Network} \cite{wang2016dueling} suggests to separate the $Q$-function estimator into two separate networks: $\hat{V}$ network to estimate the state values and $\hat{A}\text{dv}$ network to approximate the state-dependent action advantages. 

\vspace{0.2cm}
\noindent
\textbf{Example}: DQN and its variants have become the go-to RL algorithm to solve wireless problems with discrete action spaces. If the action space is continuous, researchers will propose a careful discretization scheme to fit into the Q-learning framework. As an example, \cite{wang2018DQN} solves the dynamic multichannel access problem in wireless networks using DQN. Table \ref{tab:SOTA} lists different references that used DQN in multi-agent scenarios.

\subsection{Deterministic Policy Gradient (DPG) Algorithms}
The $\max$ operator in (\ref{eq:qtarget}) limits the $Q$-learning algorithms to discrete action space (see Table \ref{tab:MFRL}). In fact, when the action space is discrete, it is tractable to compute the maximum of the $Q$-values. However, when the action space becomes continuous, finding the maximum involves costly optimization problem. In this context, DPG algorithms \cite{silver2014DPG} can be considered as an extension of $Q$-learning to continuous action spaces by replacing the $\max_{a^{\prime}} Q_{\phi}(s^{\prime}, a^{\prime})$ in (\ref{eq:qtarget}) by $Q_{\phi}(s^{\prime}, \mu_{\theta}(s^{\prime}))$ where $\mu_{\theta}(s^{\prime}) = \argmax_{a^{\prime}} Q_{\phi}(s^{\prime}, a^{\prime})$. Thus, DPG algorithms concurrently learn a $Q$-function $Q_\phi$ and a policy $\mu_\theta$. The $Q$-function is learned by minimizing the \emph{Bellman error} as explained in the previous section. As for the policy, the objective is to learn a \emph{deterministic policy} that outputs the action corresponding to $\max_a Q_\phi(s,a)$. The policy is called deterministic because it gives the exact action to take at each state $s$. Hence, the learning process consists in performing gradient ascent with respect to $\theta$ to solve the objective in (\ref{eq:6}):
\begin{align}\label{eq:6}
    J(\theta) &= \mathbb{E}_{s \sim D} \bigg[ Q_\phi(s,\mu_\theta(s)) \bigg]. \\
    \nabla_\theta J(\theta) &= \mathbb{E}_{s \sim D} \bigg[\nabla_{\theta} \mu_{\theta}(s) \nabla_{a} Q_\phi(s,a)|_{a=\mu_{\theta}(s)}\bigg].
\end{align}

Note that in the gradients formula above, the term $\nabla_{a} Q_\phi(s,a)$ requires a continuous action space. Therefore, one drawback of the DPG methods is that they cannot be applied to discrete action spaces. Observe as well that in (\ref{eq:6}), the state is sampled from a replay buffer $D$ which means that DPG algorithms are off-policy.  

Deep Deterministic Policy Gradient (DDPG) \cite{lillicrap2015DDPG} has been widely used to solve wireless communication problems. It is an extension of the DPG method where the policy $\mu_\theta$ and the critic $Q_{\phi}$ are both DNNs. Recently, two variants of the DDPG algorithm have been proposed: Twin Delayed DDPG (TD3) \cite{fujimoto2018TD3} addresses the overestimation problem discussed in \S~\ref{sec:QL} whereas in Soft Actor Critic (SAC) \cite{haarnoja2018SAC} the agent maximizes not only its expected return but also the policy entropy to improve robustness and stability. 
\begin{table*}
    \centering
    \caption{Action and state spaces for model-free RL algorithms}
    \label{tab:MFRL}
    \begin{tabular}{lcccccc}
        \hline
        \multirow{2}{*}{} & \multicolumn{2}{c}{Action} & \multicolumn{2}{c}{State} & On policy & Off policy\\
        \cline{2-5}
         & Discrete & Continuous & Discrete & Continuous \\
        \hline
        \textbf{Policy-Based} & & & \\
        REINFORCE \cite{williams1992simple}& \cmark & \cmark & \cmark & \cmark & \cmark\\
        A2C-A3C \cite{mnih2016A3C}& \cmark & \cmark & \cmark & \cmark & \cmark\\ 
        PPO \cite{schulman2017PPO}& \cmark & \cmark & \cmark & \cmark & \cmark\\ 
        TRPO \cite{schulman2015TRPO}& \cmark & \cmark & \cmark & \cmark & \cmark\\
        \textbf{Value-Based} & & & \\
        DQN/DDQN \cite{mnih2013DQN}&  \cmark & \xmark & \cmark & \cmark & & \cmark\\ 
        Dueling DQN \cite{wang2016dueling}& \cmark & \xmark & \cmark & \cmark &&\cmark\\
        \textbf{Deterministic PG} & & & \\
        DDPG \cite{lillicrap2015DDPG}&  \xmark & \cmark & \cmark & \cmark &&\cmark\\ 
        TD3 \cite{fujimoto2018TD3}&  \xmark & \cmark & \cmark & \cmark && \cmark\\
        SAC \cite{haarnoja2018SAC}&  \xmark & \cmark & \cmark & \cmark && \cmark\\
        \hline 
    \end{tabular}
\end{table*}

\vspace{0.2cm}
\noindent
\textbf{Example}: DDPG has been widely used in continuous wireless problems. In \cite{qiu2019DDPG}, DDPG is applied in energy harvesting wireless communications. Similarly, Table \ref{tab:SOTA} lists several references using DDPG agents to solve multi-agent wireless problems such as computation offloading, edge caching, UAV management, etc. 

\subsection{Theoretical Analysis and Challenges of RL}
After reviewing the key MFRL algorithms, we will discuss selective theoretical problems in both policy-based and value-based methods. First, we will study the instability problems in value-based methods with function approximators. The stability means that the error $\hat{Q}-Q^*$ gets smaller when the number of iterations increases. As mentioned in the previous section, the stability of the tabular $Q$-learning algorithm is guaranteed thanks to the contraction property of the Bellman operator. However, the contraction property is not satisfied when the Bellman error is minimized. 
Next, we focus on the convergence rate and sample complexity of policy-based methods. Sample complexity is defined as the minimal number of samples or transitions to estimate $Q^*$ and achieve a near-optimal policy, and the convergence rate determines how fast the learned policy converges to the optimal solution. Finally, we examine the interpretability issue of DRL methods which is a crucial challenge towards the application of DRL in real-world problems.

\vspace{0.2cm}
\paragraph{\textbf{Instability of off-policy TD learning with function approximators}}

In tabular RL, value-based RL methods are guaranteed to converge to the optimal value function which is the fixed point of the Bellman optimality equation. This fundamental result is justified by the contraction property of the Bellman operator. Hence, successive applications of the Bellman operator converge to the unique fixed point. However, these methods combined with function approximators suffer from instability and divergence. This is commonly referred to as the \emph{deadly triad}: {\em function approximation}, {\em bootstrapping}, and {\em off-policy training} \cite{van2018deep}. These three elements are important to consider in any RL method. Function approximation is crucial to handle large state spaces. Bootstrapping is known to learn faster than MC methods \cite{sutton2018reinforcement} thus this data efficiency is an advantage from using bootstrapping. Finally, off-policy learning enables exploration since the behavior policy is often more exploratory than the target policy. Therefore, trading off one of these techniques means losing either generalization power, data efficiency, or exploration. This is why several research efforts are dedicated to finding stable and convergent algorithms for off-policy learning with (nonlinear) function approximators and bootstrapping. 

Fixed-point methods rely on reducing the Bellman error to learn an approximation of the optimal state-value or state-action functions. As a reminder, here is the expression of the objective function governed by the Bellman error:
\begin{equation*}
\resizebox{.98\hsize}{!}{
$L(\hat{Q}) = \mathbb{E}_{s,a}\Bigg[ \bigg(R(s, a) + \gamma \mathbb{E}_{s^\prime}[\max_{a^\prime} \hat{Q}(s^\prime, a^\prime)] - \hat{Q}(s,a) \bigg)^2 \Bigg]$.
}
\end{equation*}

To compute an unbiased estimate of the loss, two samples of the next state $s^\prime$
 are needed because of the inner expectation. This is a well-known problem called the \emph{double sampling} issue \cite{baird1995}. 
 The implication of this issue is a state $s$ must be visited twice to collect two independent samples of $s^{\prime}$ which is impractical. To overcome this problem, different approaches have been adopted. The first approach reformulates the Bellman error as a \emph{saddle-point} optimization problem where a new concave function $\nu: S\times A \mapsto \mathbb{R}$ is introduced such that the loss becomes $L(\hat{Q}, \nu)=2\mathbb{E}_{s,a, s^\prime}\big[\nu(s,a)\big(\hat{Q}(s,a) - R(s, a) - \gamma \max_{a^\prime} \hat{Q}(s^\prime, a^\prime) \big)\big] - \mathbb{E}_{s,a, s^\prime}\big[ \nu(s,a)^2 \big]$ \cite{dai2017Saddlepoint}. This is called a saddle-point problem since the loss will be minimized with respect to the $Q$-function parameters and maximized with respect to $\nu$ parameters. In this context, a recent work \cite{dai2018sbeed} proposed a convergent algorithm with nonlinear function approximators (e.g. NN) and off-policy data.

Another approach to tackle this issue is to replace the $Q$-function in the inner expectation term with another target function. The choice of the target function can be an old version of the $Q$-function as in the famous DQN algorithm or the minimum of two target $Q$-functions as in TD3 and SAC. This gives a theoretical explanation of the role of target networks in stabilizing $Q$-learning with DNN.

\vspace{0.2cm}
\paragraph{\textbf{Convergence of PG with neural function approximators}}
In this part, we will summarize the recent convergence results of PG algorithms with NN as function approximators. 

Due to the non-convexity of the expected cumulative rewards in (\ref{eq:PG1}) with respect to the policy and its parameters, the analysis of the global optimality of stationary points is a hard problem. Besides, the policy gradients in (\ref{eq:PG3}) are obtained using sampling and in practice, an approximate $Q$-function is learned to estimate the expected return. In finite-horizon tasks, estimating the $Q$-function using MC rollouts results in an unbiased approximation but a biased $Q$-function for discounted infinite-horizon. 

To tackle the unbiasedness issue, \cite{sutton2000PG} introduces the \emph{compatible function approximation} theorem requiring the approximate $Q$-function $\text{Q}_\phi$ should satisfy two conditions: (i) \emph{compatibility} condition with the policy $\pi_\theta$ given by $\nabla_\phi \text{Q}_\phi(s,a) = \nabla_\theta \log \pi_\theta(a|s)$, (ii) $\text{Q}_\phi$ is learned to minimize $\mathbb{E}_{\pi_\theta} [1/2 (Q^{\pi_\theta}(s,a) - Q_{\phi}(s,a))^2]$. If these two conditions are verified, the estimates of the policy gradients are unbiased. Another approach presented in \cite{zhang2019PGTherory} is called \emph{random-horizon PG} and proposed to use rollouts with random geometric time horizons to unbiasedly estimate the $Q$-function for the infinite-horizon setting. 

Armed with the advances in non-convex optimization, several research efforts (\cite{papini2018SVRG}, \cite{shen2019PGTheory}, \cite{xu2020SVRG}, \cite{zhang2019PGTherory}) propose variants of the REINFORCE algorithm with a rate of convergence to first- or second-order stationary points. \cite{liu2019PPOConvergence} studies the non-asymptotic global convergence rate of PPO and TRPO algorithms parametrized with neural networks. These methods converge to the optimal policies at the rate of $O(\sqrt{1/T})$, where $T$ is the number of iterations. In addition, the work in \cite{wang2019ACConvergence} establishes the global optimality and convergence rate of (natural) actor-critic methods where both the actor and the critic are represented by a NN. A rate of $O(\sqrt{1/T})$ is also proved and the authors emphasize the importance of the ``compatibility" condition to achieve convergence and global optimality. In \cite{xu2020ACConvergence}, better bounds on sample complexity of (natural) actor-critic are provided. In fact, the authors demonstrate that the overall sample complexity for the mini-batch actor-critic algorithm to reach an $\epsilon$-accurate stationary point is of $O(\epsilon^{-2}\log(1/\epsilon))$ and that the natural actor-critic method requires $O(\epsilon^{-2}\log(1/\epsilon))$ samples to attain an $\epsilon$-accurate globally optimal point. In the same vein, the works in \cite{wu2020Two-timeAC} and \cite{xu2020Two-timeAC} establish the convergence rates and sample complexity bounds for the two-time scale scheme of (natural) actor-critic methods where the actor and the critic are updated simultaneously with different learning rates. Furthermore, \cite{yu2019cPGSafeRL} studies the convergence of PG in the context of constrained RL where the objective and the constraints are both non-convex functions.

\vspace{0.2cm}
\paragraph{\textbf{On eXplainable RL (XRL)}}
XRL is crucial to fully trust RL algorithms to be deployed in real-world scenarios. Recently, a lot of efforts have focused on designing explainability algorithms for AI. However, most of these works are tailored to supervised learning. XRL is more challenging due to its unsupervised nature and the dependence on complex DNN to achieve good performance. We will refer to the model or decisions to be explained as \emph{explanandum} and the generated explanations as \emph{explanans}.

According to \cite{puiutta2020XRL}, the term \emph{interpretability} is defined as the ability to not only explain the model's decisions but also to present these explanations in an understandable way to offer the possibility for non-expert users to predict the model’s behavior. A common taxonomy classifies interpretability models along two main dimensions: the scope or level of the explanans (\emph{global} vs \emph{local}) and the time when the explanations are generated (\emph{intrinsic} vs \emph{post-hoc}). Global approaches explain the behavior of the whole model, whereas local ones provide explanations to local predictions. The intrinsic or transparent category encompasses models constructed to be self-explanatory by reducing their complexity. However, post-hoc interpretability algorithms seek to provide explanations of an ML model after training. Interpretability algorithms can also be either model-specific if they are restricted to a precise explanandum or model-agnostic if they are applicable for any model/task. Intrinsic interpretability is hard to achieve in RL since they rely on complex DNN, thus post-hoc models are preferred. In addition, several XRL methods focus on specific tasks/architectures, which limit their applicability to different settings. Consequently, the design of a model-agnostic approach, independent of the RL environment, RL learning method, and tasks, is still an ongoing research direction. Furthermore, the current XRL algorithms provide explanations targeted to a knowledgeable audience. Designing more understandable approaches for a broader audience is also a pressing issue to overcome in XRL. We refer the reader to \cite{puiutta2020XRL}, \cite{heuillet2020XRL}, and the references therein for a detailed overview of explainability methods in RL. 

\section{Single Agent Model-Based RL Algorithms}
\label{sec:MBRL}
\subsection{Introduction to MBRL and Planning}

\begin{figure*}[t!]
    \begin{subfigure}[t]{0.6\textwidth}
        \includegraphics[scale=0.5]{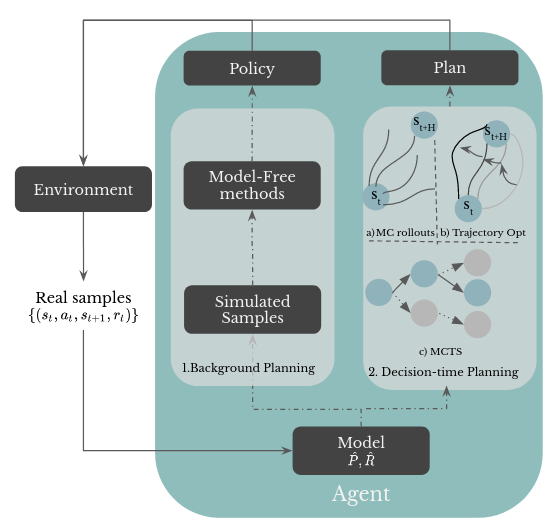}
        \caption{General Description of MBRL.}
        \label{fig:mbrl}
    \end{subfigure}%
    ~ 
    \begin{subfigure}[t]{0.39\textwidth}
        \centering
        \resizebox{0.95\textwidth}{!}{\begin{tabular}[b]{lcc}
        \specialrule{1pt}{1pt}{1pt}
        \hline
        & MF & MB  \\
        Asymptotic rewards & \cellcolor{green1!25}$+$ & \cellcolor{yellow!25}$+/-$ \\ 
        Computation at deployment & \cellcolor{green1!25}$+$ & \cellcolor{yellow!25}$+/-$  \\ 
        Data Efficiency & \cellcolor{red1!25}$-$ & \cellcolor{green1!25}$+$  \\
        Adaptation to changing rewards &  \cellcolor{red1!25}$-$ & \cellcolor{green1!25}$+$ \\ 
        Adaptation to changing dynamics &  \cellcolor{red1!25}$-$ & \cellcolor{green1!25}$+$ \\ 
        Exploration &  \cellcolor{red1!25}$-$ & \cellcolor{green1!25}$+$  \\
        \hline 
        \vspace{3cm}
    \end{tabular}}
        \caption{Model-free RL vs. MBRL.}
        \label{tab:MBRLvsMFRL}
    \end{subfigure}
    \caption{(a) Real experience from interactions with the environment is collected to learn a model. The model can be utilized in two fashions: \textbf{(1)} Background planning where simulated experience, generated by the model, is used to optimize a policy by applying any model-free learning methods on the simulated data. This planning approach is compatible with discrete and continuous action spaces; \textbf{(2)} Decision-time planning family seeks to find the best sequence of actions for the current state using the learned model. Depending on the size of the action space, different methods can be applied: For discrete action space, Monte Carlo rollouts \textbf{(2.a)} or Monte Carlo Tree Search \textbf{(2.c)} are well-known approaches. Decision-Time planning for continuous action spaces relies on trajectory optimization \textbf{(2.b)} where a random action path is chosen and updated to maximize the trajectory reward. (b) Comparative analysis of MBRL and model-free RL. The cases marked in yellow indicate that it depends on the used MBRL methods.}
\end{figure*}

Planning refers to computing an optimal policy $\pi^{*}$ assuming that the MDP $M$ is known. Classical DP methods such as Policy Iteration and Value Iteration are planning algorithms. In MBRL, the agent computes an approximate model of the environment denoted by $\hat{M}$ and performs planning in $\hat{M}$ to find the optimal actions. In contrast, model-free methods, represented previously, rely on \emph{learning} to obtain optimal policies. Therefore, MBRL has two important components for decision-making: the approximate model and the planning algorithm. First, we will present the concepts for learning world models and afterward, the planning methods are discussed. Figure (\ref{fig:mbrl}) provides a holistic illustration of MBRL framework.

The learned models are representations of knowledge about the real environment. In simpler words, the model can be an approximation of the transition function $\hat{P}(s^\prime | s, a)$ and/or the reward function $\hat{R}(s, a)$. Consequently, for each state-action pair $(s, a)$, the model predicts the next state and the immediate reward. The model can be \emph{distributional} if it outputs the distribution over all possible next states or \emph{sampled} in case only one possible next state is produced according to the computed probabilities. Furthermore, the model can be either \emph{global} or \emph{local}. Global models learn a representation of the whole state space whereas a local model is valid only in regions of the state space. Local models are more task-related and are easier to learn compared to global ones that require samples from the whole state space.
Practically, different settings are considered for learning a model. In some applications, the agent may not have access to the real state of the environment. This is the case in POMDPs where the agent only receives partial observations of the environment. 

Another consideration is the dimension of the states. It might be impossible to learn a model over high-dimensional state spaces. Hence, state representation is an important feature of the learned dynamics. For state-based models, the transition function $\hat{P}_{\theta}(s^\prime|s, a)$ is parameterized by a DNN to learn \emph{domain-agnostic} dynamics. The agent collects a dataset of interactions $D= \{(s, a, s^{\prime})\}_t$ and the network is trained using Maximum Likelihood. If the agent has access only to observations, it is possible to learn an observation transition model to directly estimate the transition probabilities between observations. In the case of high-dimensional states or observations, the common approach is to infer a latent space from the states/observations using techniques such as Variational Autoencoders (VAE) and learn a transition model is the obtained latent space. In contrast to these \emph{parametric} methods, several \emph{non-parametric} approaches can be adopted, such as episodic memory \cite{van2019MBRL}, transition graphs \cite{zhang2018transitiongraphs}, and Gaussian Processes \cite{deisenroth2011pilco}. 
 
Armed with our knowledge of methods to learn a model of the environment, we will investigate how the obtained dynamics are utilized for control. It is straightforward to observe that \emph{simulated experience} can be collected using the learned model to augment the available real-world samples. Planning is therefore performed by applying the known RL methods such as $Q$-learning to estimate value functions using the simulated and real experience. The estimated value functions are thereby utilized to optimize or improve a policy. In this context, \emph{Dyna-Q} algorithm \cite{sutton1990DynaQ} is a famous implementation of this approach of planning. It intertwines the learning and planning of the $Q$-function. The planning is performed by selecting uniformly random $K$ initial state-action pairs $(s, a)$, simulating the environment to obtain the next states and rewards, and update the $Q$-values of the sampled pairs. Selecting the starting state-action pairs in a uniformly random manner is not always efficient since all the pairs are treated equally. Different methods are proposed to address this issue. \emph{Prioritized sweeping} \cite{moore1993prioritizedsweeping} assigns priorities to state-action pairs according to the changes in their estimated values. In \emph{trajectory sampling} methods, the simulation is performed by collecting trajectories by following the current policy to update the $Q$-values. All of these methods fall into the realm of \emph{background planning} where the learned model is involved indirectly in the computation of the optimal policies.

Another approach of planning relies on the learned model to select actions at the current state $s$. This is called \emph{decision-time planning} \cite{sutton2018reinforcement}. In this category, neither a policy nor a value function is required to act in the environment. In fact, action selection is formulated as an optimization problem (in (\ref{eq:planning})) where the agent chooses a sequence of actions or a plan that maximizes the expected rewards over a trajectory of length $H$:
\begin{equation}
\label{eq:planning}
    a_1, ...., a_H = \argmax_{a_1,....,a_H} \mathbb{E} \bigg[\sum_{t=1}^H R(s_t,a_t) | a_1, ...., a_H\bigg].
\end{equation}

There are two approaches to solve this planning problem according to the size of the action space. For discrete action spaces, decision-time planning encompasses \emph{heuristic search}, \emph{MC rollouts}, and \emph{Monte Carlo Tree Search}. Heuristic search consists in considering a tree of possible continuations from a state $s$ and the values of the next actions are estimated recursively by backing up the values from the leaf nodes. Then, the action with the highest value is selected. MC rollout planning uses the approximated model to generate simulated trajectories starting from each possible next action to estimate its value (see Figure \ref{fig:mbrl}.2.a). At the end of this process, the action with the highest value is chosen in the current state. The MCTS algorithm keeps track of the expected return of state-action pairs encountered during MC rollouts to direct the search toward more rewarding pairs. At the current state, MCTS expands a tree by executing an action according to the estimated values. In the next state, MCTS evaluates the value of the obtained state by simulating MC rollouts and propagates it backward to the parent nodes. The same process is repeated in the next state. Two important observations can be made regarding these approaches: (i) the estimated state-action values are completely discarded after the action is selected. None of these algorithms stores $Q$-functions. And (ii) all these approaches are \emph{gradient-free}.

The continuous action setting is more involved because it is complicated to perform a tree search. Alternatively, \emph{trajectory optimization} methods consider one possible action sequence $a=\{a_1, \dots, a_H\}$ sampled from a fixed distribution. This sequence is then executed and the trajectory return $J(a) = \sum_{t=0}^{H} R(s_t, a_t)$ is computed. Notice that the return is a function of the action sequence. If the learned model is differentiable, it is possible to compute the gradients of $J(a)$ with respect to the actions and update the action sequence accordingly (i.e. $a = a + \nabla_a J(a)$). This planning algorithm, known as the random \emph{shooting} method, exhibits several caveats such as sensitivity to the initial action selection and poor convergence guarantees. This motivated a lot of varieties of planning methods for continuous action spaces. For example, \emph{Cross Entropy Method} (CEM) is a famous planning approach to escape local optima which shooting methods suffer from. Compared to the shooting method, the main idea of CEM is to consider a normal distribution with parametrized mean and covariance. The trajectories are sampled around the mean and the average reward is computed for each sample to evaluate its fitness. Afterward, the mean and the covariance are updated using the best samples. CEM is a simple and gradient-free approach and can exhibit fast convergence.
For a detailed comparison and benchmark of MBRL methods, we refer the interested reader to \cite{MBRLBench}. 


In Table \ref{tab:MBRLvsMFRL}, a comparison between MFRL and MBRL methods, inspired from \cite{mbrltutoicml2020}, is provided. Although model-free methods can exhibit better asymptotic reward performance and are more computationally efficient for deployment, model-based algorithms are more data efficient, robust to changes in the environment dynamics and rewards, and support richer forms of exploration. MBRL is preferred to model-free RL in multi-task settings. In fact, the same learned dynamics can be used to perform multiple tasks without further training.

However, several practical considerations should be considered for MBRL. Since the model is learned via interactions with the environment, it is prone to the following problems: (1) insufficient experience, (2) function approximation error, (3) small model error propagation, (4) error exploitation by the planner, and (5) less reliability for longer model rollouts \cite{MBRLTrust}. To avoid these problems, it is recommended to continuously re-plan to avoid error accumulation. Limited data can also cause model uncertainty which can be reduced by observing more data. Thus, it is better to estimate the uncertainty in the model predictions to know when to trust the model to generate reliable plans/actions.  One approach to estimate model uncertainty is Bayesian where methods such as Gaussian Processes \cite{deisenroth2011pilco} or Bayesian NN (i.e. \cite{gal2016PILCOBNN}) are applied. The work in \cite{chua2018boostrapEnsemble} proposed to use ensemble methods (bootstrapping) for uncertainty estimation where an ensemble of models is learned and the final predictions are the combination across the models' predictions.

\subsection{Applications of MBRL}
MBRL approaches received less interest from the wireless communication community compared to their model-free counterparts. However, we argue that MBRL is important to build practical systems. For instance, since it is hard to collect data from a real wireless network, models are trained using a simulator. The most important barrier to deploying such models in real-world settings is the \emph{reality gap} where the learned policy in a simulator does not perform as good in the real world. This is a serious issue in the context of building RL-based algorithms for 6G networks. This line of work is called \emph{sim2real} which is an active area of research in robotics. In this context, the learned models are means to bridge the gap between the simulation and the real world.  Furthermore, MBRL is advantageous because a learned model for a source task can be re-used to learn a new task faster. Coupled with meta-learning techniques, MBRL is applied to generalize to new environments and changes in the agent's world. As an example, aerial networks or drone networks,  a key enabler of 6G systems, can benefit from MBRL for a wide variety of applications such as hovering and maneuvering tasks \cite{polydoros2017survey}. Another potential application of MBRL is related to task offloading in MEC. A model can be learned to predict the load levels in edge servers. This will help the edge users to make more efficient offloading decisions, especially for delay-sensitive tasks. The main challenge in applying MBRL in multi-agent problems is the non-stationarity issue. One key application of MBRL in multi-agent problems is \emph{opponent modeling}. It consists in learning models to represent the behaviors of other agents. In multi-agent systems, \emph{opponent modeling} is useful not only to promote cooperation and coordination but also to account for the behavior of the opponents to compensate for the partial observability (See Section \ref{sec:MARLchallenges}). In addition, modeling other agents enables the decentralization of the problem because the agent can use the learned models to infer the strategies and/or rewards. 

The work in \cite{bargiacchi2020MARLMBRL} proposes a model-based algorithm for cooperative MARL. It is called \emph{Cooperative Prioritized Sweeping}. This paper extends the prioritized sweeping algorithm mentioned above to the multi-agent setting. The environment is modeled as \emph{factored MMDP} and represented by a \emph{Dynamic Bayesian Network} in which state-action interactions are represented as a coordination graph. This assumption allows the factorization of the transition function, reward function as well as the $Q$-values. Thus, these functions can be learned in an efficient and sparse manner. One drawback of this method is it assumes that the structure of the factored MDP is available beforehand which is impractical in some applications with high mobility.
The authors in \cite{krupnik2020MARLMBRL} consider two-agent competitive and cooperative settings in continuous control problems. The problem addressed in \cite{krupnik2020MARLMBRL} is how to learn separable models for each agent while capturing the interactions between them. The approach is based on \emph{multi-step generative models}. Instead of learning a model using one-step samples $(s,a,s^\prime)$, multi-step generative models utilize past trajectory segments $T_p$ with length $H$, $T_p = \{(s_{t-H-1}, a_{t-H-1}), \dots, (s_t,a_t))\}$ to learn a distribution over the future segments $T_f= \{(s_{t+1}, a_{t+1}), \dots, (s_{t+H},a_{t+H}))\}$. Hence, an \emph{encoder} will learn a distribution over a latent variable $Z$ conditioned on future trajectory segments $Q(Z | T_f)$ and a \emph{decoder} reconstructs $T_f$ such that $\hat{T_f} = D(T_p, Z)$. In the 2-agent setting, the joint probability $P(T_{fx}, T_{fy})$ where $T_{fx}$ and $T_{fy}$ are the future segments for player $x$ and $y$ respectively. The key idea is to learn two disentangled latent spaces $Z_x$ and $Z_y$. To do so, the algorithm proposed in the paper uses variational lower bound on the mutual information ~\cite{chen2016infogan}. 


\section{Cooperative Multi-Agent Reinforcement Learning}
\label{sec:MARL}

\subsection{Challenges and Implementation Schemes}


This section is dedicated to discuss the challenges that arise in multi-agent problems. Several research endeavours proposed algorithms to address these issues which led to different training schemes for cooperative agents. We start by summarizing MARL challenges that we consider as fundamental in developing systems for wireless communications.

\label{sec:MARLchallenges}
\vspace{0.1cm}
\textbf{Non-stationarity}: As mentioned before, in multi-agent environments, players update their policies \emph{concurrently}. As a consequence, the agents' rewards and state transitions depend not only on their actions but also on the actions taken by their opponents. Hence, the Markov property, stating that the reward and transition functions depend only on the previous state and the agent's action, is violated and the convergence guarantees of single agent RL are no longer valid~\cite{Hernandez-LealK17}. Due to the non-stationarity, the learning agent needs to consider the behaviour of other participants to maximize its return. One way to overcome this issue is by using a central coordinator collecting information about agents' observations and actions. In this case, standard single-agent RL methods can be applied. Since the centralized approach is not favorable, the non-stationarity challenge need to be considered in  designing decentralized MARL algorithms.

\vspace{0.1cm}
\textbf{Scalability}:
To overcome the non-stationarity problem, it is common to include information about the joint action space in the learning procedure. This will give rise to a scalability issue since the joint action space grows exponentially with the number of agents. Therefore, the use of DNNs as function approximators becomes more pressing which adds complexity to the theoretical analysis of deep MARL. For systems involving multiple agents (which are usually the cases with wireless networks with many users), scalability becomes crucial. Several research endeavors aimed to overcome this issue. One example is to learn \emph{factorized} value function with respect to the actions (see Section \ref{sec:cooperation}). 

\vspace{0.1cm}
\textbf{Partial Observability}:
In real-world scenarios, the agents seldom have access to the true state of the system. They usually receive partial observations from the environment. Partial observability coupled with non-stationarity makes MARL more challenging. In fact, as stated before, the non-stationarity issue mandates that the individual agents become aware of the other agents' policies. With only partial information available, the individual learners will struggle to overcome the non-stationarity of the system and account for the joint behavior.

\vspace{0.1cm}
\textbf{Privacy and Security}:
Since coordination may involve information sharing between agents, privacy and security concerns will arise. Shared private information (i.e. rewards) with other agents is subject to attacks and vulnerabilities. This will hinder the applicability of MARL algorithms in real-world settings. This is why fully decentralized algorithms are preferred so that all the agents keep their information private. Enormous efforts have been made for addressing the privacy and security issues in supervised learning. However, in MARL, this challenge is not extensively studied. Recently, the work in \cite{pan2019privacy} has showed that attackers can infer information about the training environment from a policy in single-agent RL.

\vspace{0.1cm}
\noindent
To promote coordination while considering the challenges discussed above, different training schemes can be adopted.
\begin{itemize}[leftmargin=*]
    \item \textbf{Fully decentralized}: A simple extension of single-agent RL to multi-agent scenarios will be IL where each agent optimizes its policy independently of the other participants. Thus, the non-stationarity problem is ignored and no coordination or cooperation is considered. This technique suffers from convergence problems \cite{tan1993}. However, it may show satisfying results in practice. In fact, recent works (see Table \ref{tab:SOA}) adopted the IL to solve several resource allocation and control problems in wireless communication networks;
    \item \textbf{Fully centralized}: This approach assume the existence of a \emph{centralized unit} that can gather information  such as actions, rewards, and observations from all the agents. This training scheme alleviates the partial observability and non-stationarity problems but it is impracticable for large scale and real-time systems;
    \item \textbf{Centralized training and decentralized execution}: CTDE assumes the existence of a centralized controller that collects additional information about the agents during training but the learned policies are decentralized and executed using the agent's local information only. CTDE is considered in several MARL algorithms since it presents a simple solution to the partial observability and non-stationarity problems while allowing the decentralization of agents' policies. 
\end{itemize}

\subsection{Algorithms and Paradigms}
Coordination solutions considered in this tutorial are categorized in two families: those based on communication and those based on learning. Emergent communication studies the learning of a communication protocol to exchange information between agents. Networked agents assume the existence of a communication structure and learn cooperative behaviors through information exchange between neighbors. The second class aims to learn cooperative behaviors without information sharing. The methods that we will present are not an exhaustive list of deep MARL since we concentrate on the key concepts that can be applied for 6G technologies. We refer the readers to \cite{hernandez2019survey}, \cite{MARLSurvey2019}, and \cite{zhang2019multiagent} for a extensive review of deep MARL literature. 

\vspace{0.2cm}
\subsubsection{\textbf{Emergent Communication}}
This an active research field where cooperative agents are allowed to communicate, for example explicitly via sending direct messages or implicitly by maintaining a shared memory. 
Deep communication problems are modeled as Dec-POMDPs where agents share a communication channel in a partially observable environment and aim to maximize their joint utility. In addition to optimizing their policies, the agents learn communication protocols to collaborate better . Direct messages can be learned concurrently with the $Q$-function. An NN is trained to output, in addition to the $Q$-values, a message to communicate to the other participants in the next round. This method involves exchanging information between all the agents which is expensive. Alternatively, \emph{memory-driven} algorithms propose to use a shared memory as a communication channel. All the agents access the shared memory before taking an action and then write a response. The advantage of this method is that the agent does not communicate with the rest of the agents directly which may eventually reduce the communication cost. Besides, the agent policy depends on its private local observations and the collective memory and not on messages from all the agents.

Integrating these methods into 6G systems requires learning cost-efficient communication due to limited resources such as bandwidth. Lately, more research endeavors have focused on learning efficient communication protocols under limited-bandwidth constraints. Precisely, methods such as pruning, attention, and gating mechanisms are applied to reduce the number of messages communicated to the agents at each control round (i.e. \cite{Jiand2018}, \cite{MaoComm2019}, \cite{wang2019CommBottleneck}). In addition to learning cost-efficient communication protocols, this field has many other open questions. For example, the robustness of the learned policies to communication errors or delays caused by noisy channels, congestion, interference, etc needs to be investigated. \cite{LoweComm2019} discusses the challenges and difficulties of learning communication in multi-agent environments.
We argue that this technique can be useful in designing intelligent 6G systems. For example, the performance of MEC or aerial communications systems can be boosted by integrating communication between agents. We strongly believe that this field can benefit from the expertise of the wireless communication community to develop more efficient communication protocols while taking into consideration the restrictions of the communication medians \cite{wang2019CommBottleneck}.  

\vspace{0.2cm}
\subsubsection{\textbf{Cooperation}}
\label{sec:cooperation}
In this section, we will overview coordination learning methods without any explicit communication. As mentioned before, training independent and hence fully-decentralized agents suffer from convergence guarantees because of the non-stationarity problem. This issue is approached with different methodologies in the deep MARL literature. The first one consists in generalizing the single-agent RL algorithms to the multi-agent setting. In particular, most of the single-agent RL algorithms such as DQN rely on experience replay buffers where state transitions are stored. In the multi-agent setting, the data stored in replay memories are obsolete because agents update their policies in parallel. Several approaches were proposed to address this problem and therefore enable the use of replay buffers to train independent learners in multi-agent environments \cite{hernandez2019survey}. Another line of work focused on training cooperative agents using the CTDE framework. For policy gradients, centralized critic(s) are learned using all agents' policies to avoid non-stationarity and variance problems and actors choose actions using local information only. This method is applied, for example, in \cite{lowe2017MADDPG} to extend the DDPG algorithm to Multi-Agent Deep Deterministic Policy Gradient (MADDPG) algorithm which can be used for systems with heterogeneous and homogeneous agents. For $Q$-learning based methods, the approach is to learn a \emph{centralized but factorized} global $Q$-function. For example, in \cite{sunehag2018VDN}, the team $Q$-function was decomposed as the sum of individual $Q$-functions whereas, in \cite{rashid2018qmix}, the authors propose to use a mixing network to combine the agents' local $Q$-functions in a non-linear way. Although these methods show promising results, they can face several challenges such as representational capacity \cite{castellini2019} and inefficient exploration \cite{mahajan2019maven}. As an alternative, \cite{gupta2017CoopPS} proposes to learn a single globally shared network that outputs different policies for homogeneous agents. 
All the presented methods above have a straightforward application in wireless communications since wireless networks are multi-agent systems by definition and coordination is crucial in such systems. In fact, MADDPG has been applied in MEC and aerial networks (i.e. UAV networks). See Section~\ref{Sec:App} for more details. 

\vspace{0.2cm}
\subsubsection{\textbf{Decentralized MARL Over Networked Agents}}
\label{sec:DMARL}
Cooperative agents, modeled using a cooperative MG, usually assume that the agents share the same reward function, thus the homogeneity of the agents. This is not the case in most wireless communication problems where agents have different preferences or reward functions. For example, MEC networks encompass several types of IoT devices.  Hence, it is important to account for the heterogeneity of the agents in the design of decentralized cooperative algorithms for 6G systems. In this context, the objective is to form a team of heterogeneous agents (i.e. with different reward functions) collaborating to maximize the \emph{team-average} reward $\bar{R} = \frac{1}{N} \sum_{i \in N} R_i$. As explained in Section~\ref{sec:background}, networked agents cooperate and make decisions using their local observations including shared information by the neighbors over a communication network. The existence of the communication network enables the collaboration between agents without the intervention of a central unit.
Let $\pi^i_{\theta_i}$ be the agent's policy parametrized as a DNN. The joint policy is given by $\pi_{\theta} = \prod_{i \in N} \pi^i_{\theta_i}(a_i|s)$ and the global $Q$-function is $Q_{\theta}$ under the joint policy $\pi_{\theta}$. To find optimal policies, the policy gradients, for each agent, can be expressed as the product of the \emph{global} $Q$-function $Q_{\theta}$ and the \emph{local} score function $\nabla_{\theta_i} \log \pi^i_{\theta_i}(a_i|s)$. However, $Q_{\theta}$ is hard to estimate knowing that the agents can only use their local information. Consequently, each agent learns a local copy $Q_{\theta_i}$ of $Q_{\theta}$. In \cite{zhang2018networked}, an actor-critic algorithm is proposed where a consensus-based approach is adopted to update the critics $Q_{\theta_i}$ as the weighted average of the local and adjacent updates. We refer the interested reader to \cite{zhang2019decentralized} for other extensions and algorithms for this framework with a theoretical analysis.

\section{Applications}
\label{Sec:App}
\subsection{MARL for MEC Systems}
Multi-access edge computing (MEC) is one of the enabling technologies for 5G and beyond 5G networks. We are witnessing a proliferation of smart devices running computationally expensive tasks such as gaming, Virtual/Augmented Reality. 
Therefore, designing efficient algorithms for MEC systems is a crucial step in providing low-latency and high-reliability services. DRL has been extensively applied to solve several problems in MEC networks including task/computation offloading (i.e. \cite{rodrigues2019MECMLSUrvey}, \cite{shakarami2020Offloading}), edge caching \cite{sheraz2020MECCachingSurvey}, network slicing, resource allocation, etc \cite{mcclellan2020Mec5G}. Recently, more interest is accorded to MARL in MEC networks to account for the distributed nature of these networks.

\textbf{Task offloading} has been studied in several works from a multi-agent perspective with a focus on decentralized execution. First, we examine the works proposing fully decentralized algorithms based on the IL framework. In \cite{ZhaoChen2018}, each mobile user is represented by a DDPG/DQN agent aiming to independently learn an optimal task offloading policy to minimize its power consumption and task buffering delays. Therefore, this paper provides a fully decentralized algorithm where users decide using their local observations (task buffer length, user SINR, CSI), the allocated power levels for local execution, and task offloading. Similarly, the approach in \cite{liu2020multiagent} is based on independent $Q$-learning where each edge user selects the transmit power, the radio access technology, and the sub-channel. The problems considered in the previous papers are formulated as MGs where the global state is the concatenation of the users' local observations and the agents act simultaneously on the system to receive independent reward signals. Thus, this formalization enables the consideration of heterogeneous users with different reward functions. Furthermore, \cite{heydari2019MecOffloading} formalizes the task offloading as a non-cooperative problem where each agent aims to minimize the task drop rate and execution delay. Each mobile user is represented as an A2C agent. An energy-aware algorithm is presented in \cite{naderializadeh2019MecEnergy} where independent DQN agents are deployed in every edge server and the servers decide which user(s) should offload their computations. However, the independent nature of these works rules out any coordination between the learning agents which may hinder the convergence of these methods in practice (see Section \ref{sec:MARLchallenges}). 

Cooperation is considered in \cite{MECMARL1} where the authors used the MADDPG algorithm to jointly optimize the multi-channel access and task offloading of MEC networks in industry 4.0. The joint optimization problem is modeled as a Dec-POMDP since the agents cannot observe the status of all the channels and is solved using the CTDE paradigm. The use of MADDPG enables the coordination between agents without any explicit communication since the critic is learned using the information from all the agents but the actors are executed in a decentralized matter. Experimental results showcased the impact of cooperation in reducing the computation delay and enhancing the channel utilization compared to the IL case. 

For \textbf{edge caching}, \cite{CoopCachingZhong2019} and \cite{CoopCachingZhang2020} propose MADDPG-like algorithms to solve the cooperative multi-agent edge caching problem. Both of these works model the cooperative edge caching as Dec-POMDP and differ in the definition of the state space and reward functions. In \cite{CoopCachingZhong2019}, the edge servers receive the same reward as the average transmission delay reduction, whereas in \cite{CoopCachingZhang2020}, the weighted sum of the local and the neighbors' hit rates is considered as a reward signal to encourage cooperation between adjacent servers. Simulation results showed that the cooperative edge caching outperforms traditional caching mechanisms such as Least Recently Used (LRU), Least Frequently Used (LFU), and First In First Out (FIFO). 

To summarize, to offer massive URLLC services, the scalability of  MEC systems is crucial. We expect to see more research efforts leveraging the deep MARL techniques to study and analyze the reliability-latency-scalability trade-off of future 6G systems. For example, applying the networked agents scheme to the above-mentioned problems is one direction to explore in future works.  

\subsection{MARL for UAV-Assisted Wireless Communications}
The application of deep MAR in UAV networks is getting more attention recently. In general, these applications involve solving cooperative tasks by a team of UAVs without the intervention of a central unit. Hence, in UAV network management, decentralized MARL algorithms are preferable in terms of communication cost and energy efficiency. The decentralized over networked agents scheme is suitable for this application if we assume that the UAVs have sufficient communication capabilities to share information with the neighbors in its sensing and coverage area. However, due to the mobility of UAVs, maintaining communication links with its neighbors to coordinate represents a considerable handicap for this paradigm. 

In \cite{UAVYang2018}, the authors study the cooperation in links discovery and selection problem. Each UAV agent $u$ perceives the local available channels and decides to establish a link with another UAV $v$ over a shared channel. Due to different factors such as UAV mobility,  wireless channel quality, and perception capabilities, each UAV $u$ has a different local set of perceived channels $C_u$ such that $C_u \cap C_v \neq \emptyset$. A link is established between two agents $u$ and $v$ if they propagate messages on the same channel simultaneously. Given the local information (i.e. state) about whether the previous message was successfully delivered, each UAV's action is a pair $(v, c_u)$ denoting by $v$ the other UAV and $c_u$ the propagation channel. Each agent receives a reward $r_u$ defined as the number of successfully sent messages over time-varying channels. The algorithm proposed in \cite{UAVYang2018} is based on independent $Q$-learning with two main modifications: \emph{fractional slicing} to deal with high dimensional and continuous action spaces and \emph{mutual sampling} to share information (state-action pairs and $Q$-function parameters) between agents to alleviate the non-stationarity issue in the fully decentralized scheme. Thus, a central coordinating unit is necessary. The problem of field coverage by a team of UAVs is addressed in \cite{UAVPham2018}. The authors formulated the problem as a MG where each agent state is defined as its position in a 3D grid. The UAVs cooperate to maximize the full coverage in an unknown field. The UAVs are assumed to be homogeneous and have the same action and state spaces. The proposed algorithm is based on $Q$-learning where a global $Q$-function is decomposed using the approximation techniques Fixed Sparse Representation (FSR) and Radial Basis Function (RBF). This decomposition technique does not allow the full decentralization of the algorithm since the basis functions depend on the joint state and action spaces. Thus, the UAVs need to share information which has an important communication cost. 

Another application of MARL in UAV networks is spectrum sharing which is analyzed in \cite{UAVShamsoshoara2019}. The UAV team is divided into two clusters: the relaying UAVs which provide relaying services for the terrestrial primary users to spectrum access for the other cluster that groups sensing UAVs transmitting data packets to a fusion center. Each UAV's action is either to join the relaying or sensing clusters. The authors proposed a distributed tabular $Q$-learning algorithm where each UAV learns a local Q function using their local states without any coordination with the other UAVs. In a more recent work \cite{UAVQie2019}, the joint optimization of multi-UAV target
assignment and path planning is solved using MARL. A team of UAVs positioned in a 2D environment aims to serve $T$ targets while minimizing the flight distance. Each UAV covers only one target without collision with threat areas and other UAVs. To enforce the collision-free constraint, a collision penalty is added to the reward thus rendering the problem a mixed RL setting with both cooperation and competition. Consequently, the MADDPG algorithm is adopted to solve the joint optimization. Furthermore, \cite{UAVCui2019} formulates resource allocation in a downlink communication network as a SG and solved it using independent $Q$-learning. The work in \cite{UAVtovizka2019} applies MARL for fleet control, particularly, aerial surveillance and base defense in a fully centralized fashion.

\subsection{MARL for Beamforming Optimization in Cell-Free MIMO and THz Networks}
DRL has been extensively applied for uplink/downlink beamforming optimization. Particularly, several works focused on beamforming computation in cell-free networks. In the fully centralized version of the cell-free architecture, all the access points are connected and coordinated through a central processing unit to serve users in their coverage area. Although the application of single-agent DRL for cell-free networks has empirically shown optimal performance, the computational complexity and the communication cost increase drastically with the number of users and access points. As a remedy to this issue, hybrid methods based on dynamic clustering and network partitioning are proposed. The core idea of these methods is to cluster users and/or access points to reduce the computational and communication costs as well as to enhance the coverage by reducing interference. As an example, in \cite{Yesser2020Beamforming}, a DDQN algorithm is implemented to perform dynamic clustering and a DDPG agent is dedicated to beamforming optimization. This joint clustering and beamforming optimization is formulated as an MDP and a central unit is used for training and execution. 
In \cite{ge2020BeamformingDist}, dynamic downlink beamforming coordination is studied in a multi-cell MISO network. The authors proposed a distributed algorithm where the base stations are allowed to share information via a limited exchange protocol. Each base station is presented as a DQN agent trying to maximize its achievable rate while minimizing the interference with the neighboring agents. The use of DQN networks required the discretization of the action space which is continuous by definition. The same framework can be applied with PG or DDPG methods to handle continuous action space. 

Furthermore, THz communication channels are characterized by high attenuation and path loss which require transmitting highly directive beams to minimize the signal power propagating in directions other than the transmission direction. In this context, directional beamforming and beam selection are possible solutions to enhance the communication range and reduce interference. Intelligent beamforming in THz MIMO systems is another promising application of MARL for future 6G networks.


\subsection{Spectrum Management}
In \cite{zia2019D2D} and \cite{li2019D2D}, the spectrum allocation in Device-to-Device (D2D) enabled 5G HetNets is considered. The D2D transmitters aim to select the available spectrum resources with minimal interference to ensure the minimum performance requirements for the cellular users. The authors in \cite{zia2019D2D} consider a non-cooperative scenario where the agents independently maximize their throughput. Based on the selected resource block in the previous time step, the D2D users choose the resource block for uplink transmission and receive a positive reward equivalent to the capacity of the D2D user if the cellular users' constraints are satisfied, otherwise, a penalty is imposed. The problem is solved using a tabular $Q$-learning algorithm where each D2D agent learns a local $Q$-function with local state information. This work does not scale for high-dimensional state spaces since it is based on a tabular approach. This problem is addressed in \cite{li2019D2D} where actor-critic based algorithms are proposed for spectrum sharing. Two approaches are used to promote cooperation between D2D users. The first is called multi-agent actor-critic where each agent learns a $Q$-function in a centralized manner using the information from all the other agents. The learned policies are executed in a decentralized fashion since the actor relies on local information only. The second approach proposes to use information from neighboring agents only to train the critic instead of the information from all the agents to reduce the computational complexity for large-scale networks. The action selection and the reward function are similar to the ones defined in the previous work. In this work, the state space is richer and contains information about (i) the instant channel information of the D2D corresponding link, (ii) the channel information of the cellular link (e.g. from the BS to the D2D transmitter), (iii) the previous interference to the link, and (iv) the resource block selected by the D2D link in the previous time slot. 

Another application of MARL is resource allocation in cellular-based vehicular communication networks. The Vehicle-to-Vehicle (V2V) transmitters jointly select the communication channel and the power level for transmission. The work in \cite{ye2019V2V} proposes a fully decentralized algorithm based on DQN to maximize the Vehicle-to-Infrastructure capacity under V2V latency constraints. Although the solution is decentralized in the sense that each agent trains a $Q$-network locally, the state contains information about the other participants. The authors include a penalty in the reward function to account for the latency constraints in V2V communications. For more references on MARL in vehicular communications, we refer the reader to \cite{althamary2019surveyVehicular}.

Interference mitigation will be a pressing issue in THz communications. Exploiting the THz bands is one key enabler of 6G systems for higher data rates. In \cite{barazideh2020THZ}, a multi-arm bandit based algorithm is proposed for intermittent interference mitigation from directional links in two-tier HetNets. However, the proposed solution is valid for a single target receiver. Another work in \cite{singh2020D2DTHZ} proposes a two-layered distributed D2D model, where MARL is applied to maximize user coverage in dense-indoor environments with limited resources (i.e. a single THz access point, limited bandwidth, and limited antenna gain). Devices in the first layer can directly access the network resources and act as relays for the second layer devices. The objective is to find the optimal D2D links between the two layers. The devices from the first layer are modeled as $Q$-learning agents and decide, by using local information, the D2D links to establish with the second layer devices. The agents receive two types of rewards: a private one for serving a device in the second layer and a public reward regarding the system throughput. To promote coordination, the agents receive information about the number of their neighbors and their states. \cite{temesgene2020SecMgnt} studies a two-tier network with virtualized small cells powered by energy harvesters and equipped with rechargeable batteries. These cells can decide to offload baseband processes to a grid-connected edge server. MARL is applied to minimize grid energy consumption and traffic drop rate. The agents collaborate via exchanging information regarding their battery state. \cite{sana2020mmWave} inspects the problem of user association in dynamic mmWave networks where users are represented as DQN agents independently optimizing their policies using their local information.

\subsection{Intelligent Reflecting Surfaces (IRS)-Aided Wireless Communications}
Intelligent Reflecting Surfaces (IRS)-aided wireless communications have attracted increasing interest due to the coverage and spectral efficiency gains they provide. Multiple research works proposed DRL-based algorithms for joint beamforming and phase shift computation. These contributions study systems with a single IRS which is far from the real-world case. More recent research endeavors seek to remedy this shortcoming. For example, in \cite{he2020IRS}, the authors consider a communication system with multiple IRS cooperating together under the coordination of an IRS controller. The joint beamforming and phase shift optimization problem is decoupled and solved in an alternating manner using fractional programming.
Another line of work aims to provide secure and anti-jamming wireless communications by adjusting the IRS elements. This problem was also approached using single-agent RL (i.e. in \cite{yang2020IRSsecurejam}). For distributed deployment of multiple IRS, secure beamforming is solved in \cite{xiu2020IRSsecure} using alternating optimization scheme based on successive convex approximation and manifold optimization techniques.
To the date of the writing of this tutorial, we are unable to find a decentralized algorithm for a multi-IRS system based on MARL techniques. This is why we believe that this is a promising direction to propose distributed deployment schemes for multi-IRS systems.

\begin{table*}[ht!]
    \centering
    \caption{These papers applied MARL techniques for wireless communication problems. Learning type is the MARL training technique.}
    \label{tab:SOA}
    \begin{tabularx}{\textwidth}{c X c c c}
    \specialrule{1pt}{1pt}{1pt}
    \multicolumn{1}{c}{Work} & \multicolumn{1}{c}{Summary} & \multicolumn{1}{c}{Learning} & \multicolumn{1}{c}{Algorithm} & \multicolumn{1}{c}{Cooperation?} \\
    \hline
    \cite{ZhaoChen2018} & Proposes a decentralized computation offloading algorithm for multi-user MEC networks to allocate power levels for local execution and task offloading & IL & DDPG/DQN & \xmark \\ 
    \cite{liu2020multiagent} & Formulates computation offloading in multi-user MEC systems as MARL problem to allocate transmit power, radio access technology and subchannel & IL & $Q$-learning & \xmark\\
    \cite{heydari2019MecOffloading} & Studies the task offloading problem in a non-cooperative manner where agents learn independent policies to minimize the task drop rate and the execution
delay & IL & A2C & \xmark \\
    \cite{naderializadeh2019MecEnergy} & An energy-aware multi-agent algorithm is proposed where each edge server decide which users should offload their computations from a pool of edge devices & IL & DQN & \xmark \\
    \cite{MECMARL1} & Considers joint multi-channel access and task offloading in cooperative MEC networks& CTDE & MADDPG & \cmark \\
    \cite{CoopCachingZhong2019} & Solves cooperative edge caching in MEC systems using MARL with shared reward signal & CTDE & MADDPG-like & \cmark \\
    \cite{CoopCachingZhang2020} & Proposes different formulation (i.e. state space, reward function) of the MARL system for cooperative edge caching & CTDE & MADDPG-like & \cmark \\
    \cite{UAVYang2018} & Introduces fractional slicing and mutual sampling to learn cooperative links discovery and selection using independent $Q$-learning algorithm & IL & DQN & \cmark \\
    \cite{UAVPham2018} & Studies a team of UAVs providing the full coverage of an unknown field while minimizing the overlapping UAVs' field of views & Centralized & $Q$-learning & \cmark \\
    \cite{UAVShamsoshoara2019} & Considers spectrum sharing in UAV network and propose a distributed algorithm where each UAV decides whether to join a relaying or sensing clusters & IL & $Q$-learning & \xmark \\
    \cite{UAVQie2019} & The joint optimization of task-assignment and path planning in a multi-UAV network is studied & CTDE & MADDPG & \cmark \\
    \cite{UAVCui2019} & Solves the resource allocation problem in a multi-UAV downlink communication network using MARL & IL & $Q$-learning & \xmark \\
    \cite{UAVtovizka2019} & Develops a multi-UAV fleet control systems particularly for aerial surveillance and base defense & Centralized & PG & \cmark \\
    \cite{ge2020BeamformingDist} & Considers dynamic downlink beanmforming coordination where base stations cooperate to maximize their capacity while mitigating inter-cell interference via a limited-information sharing protocol & \textcolor{black}{Networked Agents} & DQN & \cmark \\ 
    \cite{zia2019D2D} & Considers non-cooperative spectrum allocation in D2D HetNets with stochastic geometry & IL & $Q$-learning & \xmark \\
    \cite{li2019D2D} & Formulates a cooperative spectrum sharing problem in D2D underlay communications as a decentralized multi-agent system & CTDE & MADDPG-like & \cmark \\
    \cite{ye2019V2V} & Proposes a decentralized algorithm for joint sub-band and power level allocation in V2V enabled cellular networks & IL & DQN & \xmark\\
    \cite{singh2020D2DTHZ} & A 2-layered multi-agent D2D model in THz communications is proposed to maximize user coverage in dense-indoor environments  & \textcolor{black}{Networked Agents} & Q-learning & \cmark \\
    \cite{temesgene2020SecMgnt} & Applies MARL to minimize grid energy consumption and traffic drop rate in two-tier network with virtualized small cells & \textcolor{black}{Networked Agents} & Q-learning & \cmark \\
    \cite{sana2020mmWave} & Solves user association problem in dynamic mmWave networks & IL & DQN & \xmark \\
    \hline
    \end{tabularx}
\end{table*}


\section{Conclusion and Future Research Directions}
\label{Sec:Conclusion}
We have presented an overview of model-free, model-based single-agent RL, and cooperative MARL frameworks and algorithms. We have provided the mathematical backgrounds of the key frameworks for both single-agent and multi-agent RL. Afterward, we have developed an understanding of the state-of-the-art algorithms in the studied fields. We have discussed the use of model-based methods for solving wireless communication problems. Focusing on cooperative MARL, we have outlined different methods to establish coordination between agents. To showcase how these methods can be applied for wireless communication systems, we have reviewed the recent research contributions to adopt a multi-agent perspective in solving communication and networking problems. These problems involve AI-enable MEC systems, intelligent control and management of UAV networks, distributed beamforming for cell-free MIMO systems, cooperative spectrum sharing, THz communications, and IRS deployment. We have chosen to focus on cooperative MARL applications because several surveys on single-agent RL exist in the literature. Our objective has been to highlight the potential of these DRL methods, specifically MARL, in building self-organizing, reliable, and scalable systems for future wireless generations. In what follows, we discuss research directions to enrich and bridge the gap between both fields. 
\begin{itemize}
\item \textbf{Network topologies}: 
One of the shortcomings of the MG-based methods of MARL problems is the assumed homogeneity of the studied systems. However, this is seldom the case in real-world scenarios like MEC-IoT or sensing networks. For this reason, we have motivated the networked MARL paradigm where agents with different reward functions can cooperate.
Accounting for the heterogeneity of the wireless communications systems is mandatory for practical algorithm design. Mobility is also a  challenge in wireless communication problems. Developing MARL algorithms with mobility considerations is an interesting research direction.

\item \textbf{Constrained/safe RL}: RL is based on maximizing the reward feedback. The reward function is designed by human experts to guide the agent policy search but reward design is often challenging and can lead to unintended behavior. Wireless communication problems are often formulated as optimization problems under constraints. To account for those constraints, most of the recent works adopt a reward shaping strategy where penalties are added to the reward function for violating the defined constraints. In addition, reward shaping does not ensure that the exploration during the training is constraint-satisfying. This motivates the constrained RL framework.
It enables the development of more reliable algorithms ensuring that the learned policies satisfy reasonable service quality or/and respect system constraints.

\item \textbf{Theoretical guarantees}: Despite the abundance of the experimental works around RL methods, their convergence properties are still an active research area. Several endeavors, reviewed above, proposed convergence guarantees for the policy gradient algorithms under specific assumptions such as unbiased gradient estimates. More urging theoretical questions need to be addressed such as the convergence speed to a globally optimal solution, their robustness due to approximation errors, their behaviors when limited sample data is available, etc.

\item \textbf{Privacy}: One of the challenges of the commercialization of DRL-based solutions is privacy. These concerns are rooted in the data required to train RL agents such as actions and rewards. Consequently, information about the environment dynamics, the reward functions can be inferred by malicious agents \cite{pan2019privacy}. Privacy-preserving algorithms are attracting more attention and interest. \emph{Differential privacy} was investigated in the context of Federated Learning as well as DRL \cite{wang2019privacy}. Privacy is not sufficiently explored in the context of wireless communication.

\item \textbf{Security and robustness}: DNNs are known to be vulnerable to adversarial attacks and several recent works demonstrated the vulnerability of DRL to such attacks as well. To completely trust DRL-based methods in real-world critical applications, an understanding of the vulnerabilities of these methods and addressing them is a central concern in the deployment of AI-empowered systems \cite{lin2017adversarial}, \cite{gleave2019adversarial}. In addition to adversarial attacks, the robustness of the learned policies to differences in simulation and real-world settings need to be addressed and studied (i.e. \cite{pinto2017robustRL}). 
\end{itemize}
\bibliographystyle{ieeetr}
\bibliography{references}
\end{document}